\documentclass{article}

\usepackage[letterpaper,textwidth=122mm,textheight=193mm,left=47.4mm,top=32mm,headsep=16pt]{geometry}

\usepackage{graphicx}
\usepackage{algorithm}
\usepackage{algorithmic}
\usepackage{booktabs}
\usepackage{makecell}
\usepackage[normalem]{ulem} 
\usepackage{xcolor}
\usepackage{amsmath,amssymb}

\providecommand{\keywords}[1]{\par\smallskip\noindent\textbf{Keywords:} {\def\and{\,\textperiodcentered\,}#1}}

\usepackage{xspace}
\makeatletter
\DeclareRobustCommand\onedot{\futurelet\@let@token\@onedot}
\def\@onedot{\ifx\@let@token.\else.\null\fi\xspace}

\makeatother

\usepackage[accsupp]{axessibility}  

\usepackage{hyperref}

\title{SHIFT: Motion Alignment in Video Diffusion Models with Adversarial Hybrid Fine-Tuning}

\author{
  Xi Ye$^{1,2,\dagger}$ \quad
  Wenjia Yang$^{1,\dagger}$ \quad
  Yangyang Xu$^{1,2}$ \quad
  Xiaoyang Liu$^{3}$ \\[4pt]
  Duo Su$^{1}$ \quad
  Mengfei Xia$^{2}$ \quad
  Jun Zhu$^{1*}$ \\[8pt]
  $^{1}$Tsinghua University \quad $^{2}$Ant Group \\[2pt]
  $^{3}$University of Chinese Academy of Sciences
}

\date{}

\begin{document}

\maketitle

\makeatletter
\begingroup
\renewcommand{\thefootnote}{}%
\long\def\@makefntext#1{\noindent#1}
\footnotetext{$^{\dagger}$Equal contribution.}%
\footnotetext{Accepted by ECCV 2026.}%
\endgroup
\makeatother

\begin{abstract}

Image-conditioned video diffusion models achieve impressive visual realism but often suffer from weakened motion fidelity, e.g., reduced motion dynamics or degraded long-term temporal coherence, especially after fine-tuning. We study motion alignment in video diffusion models post-training. To address this, we introduce pixel-motion rewards based on pixel flux dynamics, capturing both instantaneous and long-term motion consistency. We further propose \underline{S}mooth \underline{H}ybr\underline{i}d \underline{F}ine-\underline{t}uning (SHIFT), a scalable reward-driven framework that unifies supervised fine-tuning and advantage-weighted fine-tuning. Benefiting from novel adversarial advantages, SHIFT improves convergence speed and mitigates reward hacking. Experiments show that our approach efficiently resolves dynamic-degree collapse in modern video diffusion models supervised fine-tuning. Project page: \url{https://xiye20.github.io/projects/SHIFT/}.
\end{abstract}

\section{Introduction}
\label{sec:intro}


Recent advances in diffusion-based generative modeling have enabled high-quality video synthesis across image-to-video (I2V) and text-image-to-video (TI2V)~\cite{wan2025wan,blattmann2023stablevideodiffusionscaling,ye2022vptr}. By extending powerful image diffusion backbones with temporal modeling, modern video diffusion models produce visually compelling results with strong semantic fidelity. Yet high visual quality does not guarantee temporally faithful motion, and generated videos often lack rich dynamics.

Motion is fundamental to video realism. Beyond frame-level visual quality, realistic video generation requires coherent object trajectories, sustained motion magnitude, and physically plausible temporal evolution. However, many video diffusion models exhibit motion attenuation, where object movement becomes overly smooth, damped, or static. This issue is particularly pronounced after supervised fine-tuning (SFT) of image-conditioned models, I2V or TI2V models often suffer from a dynamic-degree degradation phenomenon: motion strength and long-term dynamics decrease even as visual sharpness and semantic alignment improve \cite{yuan2024instructvideo, zhao2024identifying, tian2025extrapolating}. Recent benchmarking further suggests that visual realism does not imply physical understanding, and that current generative video models frequently violate basic physical principles \cite{motamed2025generative}.

Reinforcement learning (RL) is a typical post-training method to improve desired quality of video diffusion models. However, aligning video diffusion models with motion-centric objective presents several challenges. First, motion is a high-dimensional, low-level, and continuous signal over spatiotemporal space, making it difficult to supervise using semantic labels. Human feedback is poorly suited for evaluating subtle physical dynamics at scale, while existing vision-language models lack the precision required for pixel-level motion assessment. Second, motion quality depends on both short-term coherence and long-term temporal structure, requiring reward signals that operate across multiple time scales.

In this work, we argue that effective motion alignment should be grounded in low-level motion representations rather than high-level semantics. We propose to model pixel trajectories via a transport residual derived from the optical-flow equation, providing a low-level, motion-aware description of temporal dynamics. Based on this representation, we design two complementary motion rewards: an Instantaneous Motion reward (IMR) that enforces local temporal coherence between adjacent frames, and a Long-term Motion reward (LMR) that captures sustained dynamics and global motion structure over extended horizons. These rewards are dense, scalable, and directly tied to pixel-level motion, enabling fine-grained supervision without human annotations or semantic models. In this way, our pixel-motion reward model addresses the failure mode of motion degradation by explicitly incentivizing both motion magnitude and temporal coherence.

Beyond the reward design, applying RL to video diffusion models poses further challenges: MDP-based methods~\cite{black2024trainingdiffusionmodelsreinforcement} are computationally expensive, tightly coupled to specific samplers, and difficult to scale; gradient-based alternatives~\cite{prabhudesai2024video,domingoenrich2025adjointmatchingfinetuningflow} require backpropagating through the full reverse process and cannot use black-box rewards. Moreover, optimizing imperfect rewards risks reward hacking, which is amplified by the long temporal horizons of video generation.

To overcome the aforementioned challenges of RL in video diffusion models, we introduce Smooth Hybrid Fine-Tuning (SHIFT) with adversarial advantage. SHIFT dynamically fuses SFT and reward-based fine-tuning into a stable unified optimization framework. Besides, SHIFT alternates between fine-tuning the diffusion model and updating the motion reward model. By explicitly incorporating adversarial training, SHIFT accelerates convergence, and mitigates reward hacking. Furthermore, unlike reverse-process reinforcement learning methods, SHIFT operates entirely in the forward diffusion process, does not require storing denoising trajectories, and remains agnostic to the choice of diffusion sampler.

Our contributions are threefold:
\begin{itemize}
    \item We propose pixel-motion reward models that capture both instantaneous and long-term motion fidelity, enabling scalable motion supervision without human preference labels.

    \item We introduce a hybrid fine-tuning framework that unifies an offline SFT anchor with advantage-weighted updates on model rollouts in the forward process, remaining efficient and sampler-agnostic.

    \item Comprehensive evaluations and ablations demonstrating that SHIFT improves motion dynamics while preserving appearance quality.
\end{itemize}
\section{Related works}
Standard I2V/TI2V diffusion denoising objective provides no explicit incentive to preserve motion magnitude, causing supervised fine-tuning to bias models toward visually stable but physically inert solutions. Recent efforts to incorporate physics information into video generation~\cite{lin2025exploring} vary in both representation (low-level implicit vs.\ high-level symbolic) and integration strategy (conditional training vs.\ external feedback). Our motion-aware reward model falls in the low-level implicit category and serves as an external feedback signal for the SHIFT fine-tuning framework.

\textbf{Motion quality in video diffusion models.}
Image-to-video diffusion models frequently suffer dynamic-degree degradation during supervised fine-tuning, where motion strength and long-term temporal coherence deteriorate even as visual quality improves~\cite{yuan2024instructvideo, zhao2024identifying, tian2025extrapolating}. Several directions have been explored to address this. Explicit geometric supervision methods such as Track4Gen~\cite{jeong2025track4genteachingvideodiffusion} employ point-tracking losses for trajectory-level guidance but rely on high-level semantic alignment, failing to capture fine-grained physical motion laws~\cite{chen2023motionconditioneddiffusionmodelcontrollable,hu2025lamd,ye2024stdiff}. Inference-time corrections~\cite{tian2025extrapolating} amplify motion via training-free model merging but do not rectify the model's underlying bias. Preference-based approaches like DenseDPO~\cite{xu2025visionrewardfinegrainedmultidimensionalhuman,rafailov2024directpreferenceoptimizationlanguage} introduce segment-level labels, yet their reliance on human perception introduces subjectivity and favors frame visual quality over motion consistency. In contrast, our method provides precise dynamic feedback through motion-aware reward models that require no human intervention.

\textbf{Reinforcement learning in diffusion models.}
RL-based post-training alignment of diffusion models encompasses MDP-based policy gradient methods \cite{black2024trainingdiffusionmodelsreinforcement,fan2023dpokreinforcementlearningfinetuning,liu2025flowgrpotrainingflowmatching}, reward-weighted regression (RWR) \cite{nair2021awacacceleratingonlinereinforcement,lee2023aligningtexttoimagemodelsusing}, and gradient-based techniques \cite{prabhudesai2024video,domingoenrich2025adjointmatchingfinetuningflow} (see supplementary material for a detailed survey). Scaling these to video diffusion, however, remains challenging: MDP-based methods incur prohibitive memory costs from storing full denoising trajectories, both MDP and RWR approaches are tightly coupled to specific samplers, e.g., SDE sampler, and long temporal horizons exacerbate credit assignment \cite{domingoenrich2025adjointmatchingfinetuningflow,wu2025densedpofinegrainedtemporalpreference}. These difficulties are compounded by reward hacking, where generators exploit imperfections in reward models rather than learning intended behaviors, which is especially severe in high-dimensional video generation \cite{wu2025rewarddancerewardscalingvisual,amodei2016concreteproblemsaisafety}. Existing mitigations, including KL regularization \cite{jaques2019wayoffpolicybatchdeep}, GRPO \cite{shao2024deepseekmath,liu2025flowgrpotrainingflowmatching}, and adversarial reward updates \cite{zha2025rl,ma2024videovideogenerativeadversarial}, address this partially. In contrast, SHIFT operates entirely in the forward process with online adversarial advantages, requiring neither human feedback nor specific diffusion samplers.
\section{Preliminaries: Diffusion RL and Limitations.}
\label{sec: preliminary}
The goal of reinforcement learning in diffusion is to fine-tune a pre-trained model $p_\theta$ to maximize a reward function $r(x_0)$, which reflects human preference or specific downstream objectives, and $x_0$ denotes a clean data example. Denoising Diffusion Policy Optimization (DDPO)~\cite{black2024trainingdiffusionmodelsreinforcement} formulates the reverse diffusion sampling process as a multi-step Markov Decision Process (MDP). Thus, DDPO requires an SDE-solver for sampling. The sampling process forms a trajectory $\tau = \{x_T, \dots, x_0\}$, where each denoising step $p_\theta(x_{t-1}|x_t)$ represents the policy action, and a reward is observed only at the final state $x_0$.

To ensure optimization stability, trust-region methods (e.g., PPO) typically impose a \textit{Reverse KL} constraint between the updated policy $p_\theta$ and a reference policy $p_{\text{ref}}$ (usually the pre-trained model):
\begin{equation}
\label{eq:normal_rl_optim}
    \max_\theta \mathbb{E}_{x_{0:T} \sim p_\theta}[r(x_0)] \quad \text{s.t.} \quad D_{\text{KL}}(p_\theta || p_{\text{ref}}) \le \gamma.
\end{equation}

The closed-form optimal solution to this constrained optimization problem is known to take the form of an energy-based distribution~\cite{nair2021awacacceleratingonlinereinforcement, peters2007reinforcement}:
\begin{equation}
\label{eq:rl_optimal_solution}
    p^*(x) \propto p_{\text{ref}}(x) \exp(\frac{r(x)}{\beta}),
\end{equation}
where $\beta$ is a temperature parameter balancing reward maximization and the KL constraint. Standard approaches optimize this objective via policy gradients.
The gradient of the regularized objective is derived as:
\begin{align}
\label{eq:pg_gradient}
    \nabla_\theta\mathcal{L}_{\text{RL}} &= -\mathbb{E}_{x_{0:T}}\left[\nabla_\theta\log p_\theta(x_{0:T})\cdot r(x_0)\right] + \nabla_\theta D_{\text{KL}}(p_\theta||p_{\text{ref}}) \\
    &= -\mathbb{E}_{x_{0:T}}[\sum_{t=1}^T\nabla_\theta\log p_\theta(x_{t-1}|x_t)\cdot r(x_0)] + \nabla_\theta D_{\text{KL}}(p_\theta||p_{\text{ref}}).
\end{align}

This formulation highlights the dependence on full trajectory optimization. For the KL-div term in Eq.\eqref{eq:pg_gradient}, we can derive a closed-form solution, please refer to \cite{liu2025flowgrpotrainingflowmatching} for more detail. Empirically, prior works utilize the relative group advantage $A(x_0)$ instead of raw reward $r(x_0)$ for a more stable training \cite{liu2025flowgrpotrainingflowmatching}. While effective for lightweight image models, applying PPO-style RL to large-scale video diffusion models presents significant challenges:

\textbf{1. Computational Prohibitiveness of On-Policy Sampling.} 
PPO-style updates require frequent on-policy sampling to keep the importance sampling ratio close to 1. For high-resolution video models, generating fresh rollouts is computationally prohibitive. Stale data leads to divergent importance weights and high-variance updates, creating a dilemma: we cannot afford frequent online sampling, yet cannot train stably without it.

\textbf{2. Structural Mismatch between MDP and Diffusion.}
Formulating diffusion fine-tuning as a standard MDP imposes significant solver restrictions and discretization inconsistencies. While diffusion models are trained to approximate continuous marginal vector fields (via score or flow matching) to support flexible inference, MDP-based RL methods lock the training process to specific discrete reverse chains. This introduces a "forward-reverse inconsistency" where the fine-tuning objective diverges from the pre-training paradigm \cite{chen2025bridging}. Furthermore, optimizing discrete trajectories complicates integration with inference-time techniques like Classifier-Free Guidance (CFG) \cite{chen2025bridging, zheng2025diffusionnft}.

\section{Methodology}
In this work, we fine-tune pre-trained video diffusion models to enhance \emph{motion fidelity}, implicitly defined by pixel-level motion consistency. We posit that real video distributions inherently exhibit correct motion dynamics. To align the generated distribution with this prior, we introduce a motion-aware reward mechanism based on optical-flow features, operating across both instantaneous and long-term temporal scales, see Figure~\ref{fig:reward-model}(a).

We then propose \textbf{SHIFT} (Smooth Hybrid Fine-tuning, Figure~\ref{fig:reward-model}(b)), a framework that fuses Advantage-Weighted Regression (AWR) on generated samples with an SFT anchor on real data to ensure stability. To mitigate reward hacking, we adopt an adversarial paradigm, alternating between diffusion model updates and reward model refinement.

\subsection{Pixel-Motion Reward Models}
\label{sec:reward}
\begin{figure}[t]
  \centering
  \begin{minipage}[t]{0.48\linewidth}
    \centering
    \includegraphics[width=\linewidth]{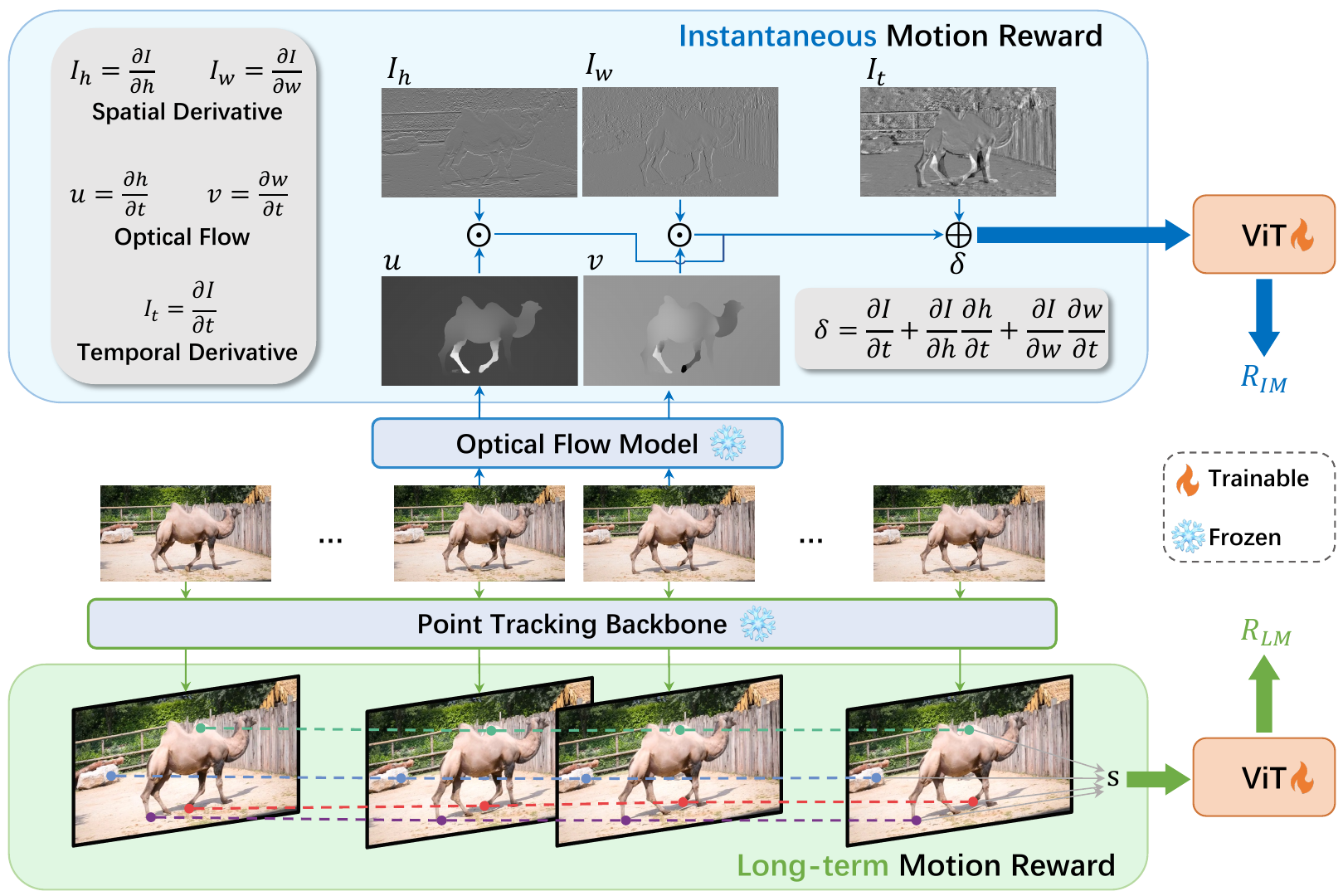}
    \centerline{\small (a) Reward models}
  \end{minipage}
  \hfill
  \begin{minipage}[t]{0.48\linewidth}
    \centering
    \includegraphics[width=\linewidth,clip,trim=190 75 200 55]{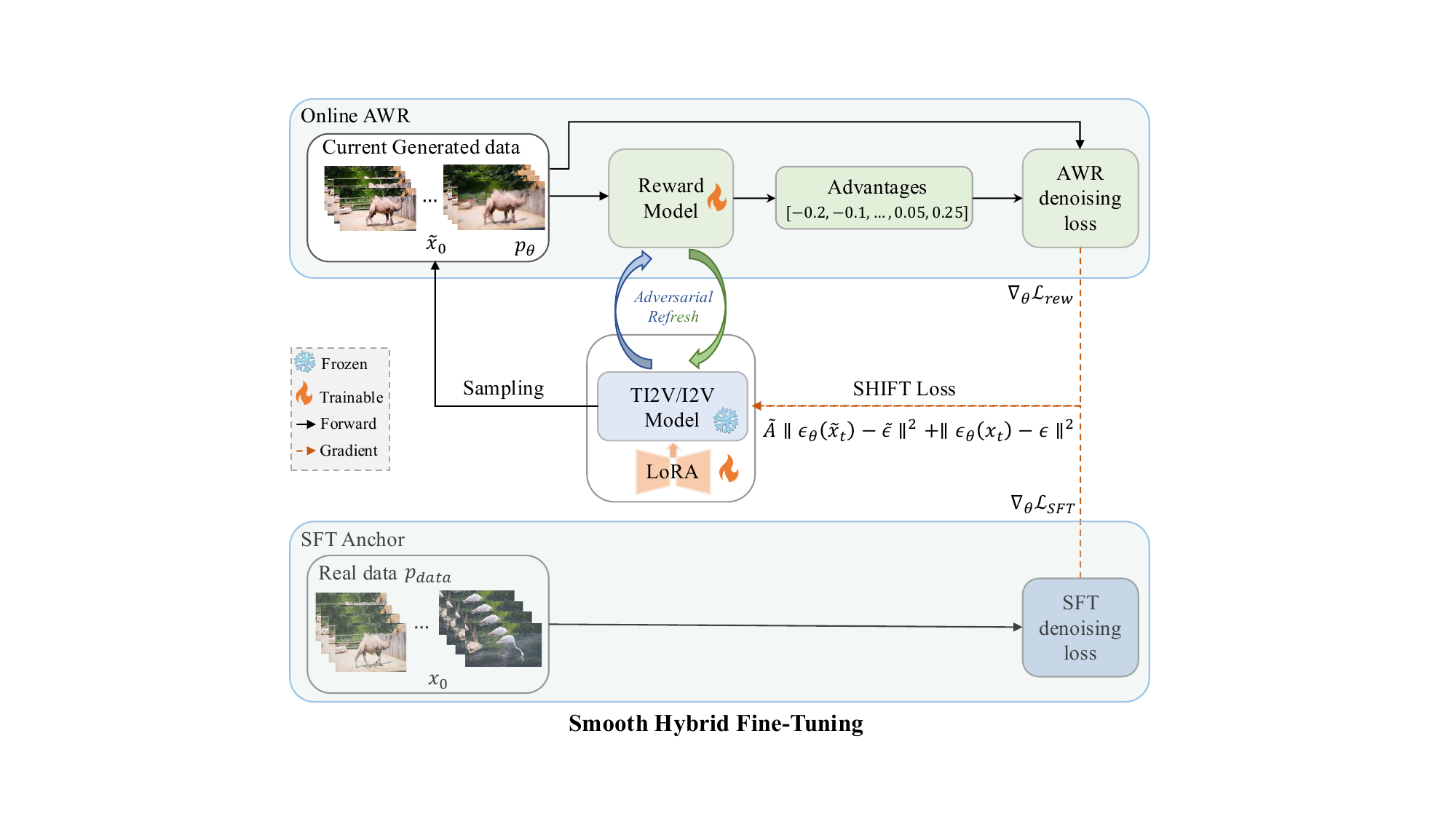}
    \centerline{\small (b) Smooth Hybrid Fine-Tuning framework}
  \end{minipage}
  \caption{Overview of the proposed approach. (a)~Pixel-motion reward models: instantaneous reward based on optical-flow residual (top) and long-term reward based on trajectory dynamics (bottom). (b)~The SHIFT fine-tuning framework.}
  \label{fig:reward-model}
  \end{figure}

\textbf{Instantaneous Motion Reward (IMR).}
To capture local temporal coherence, we model video as a continuous scalar field of pixel intensity $I(\mathbf{x}, t): \Omega \times \mathbb{R}^+ \rightarrow \mathbb{R}$, defined over the spatial image domain $\Omega \subset \mathbb{R}^2$ with coordinates $\mathbf{x} = (h, w) \in \Omega$ and time $t$. Inspired by fluid dynamics, the evolution of intensity is related to a continuity equation, which decomposes temporal changes into distinct motion components:
\begin{align}
\label{eq:pixel_flow_decomp}
  S = \frac{\partial I}{\partial t} + \nabla \cdot (I\mathbf{u}) = \frac{\partial I}{\partial t} + \underbrace{\mathbf{u} \cdot \nabla I}_{\text{Advection}} + \underbrace{I (\nabla \cdot \mathbf{u})}_{\text{Divergence}},
\end{align}
where $\mathbf{u} = (u, v) = (\frac{\partial h}{\partial t}, \frac{\partial w}{\partial t})$ represents the pixel velocity field.
The advection term, $\mathbf{u} \cdot \nabla I$, captures the transport of pixel values along motion trajectories (e.g., object translation). The divergence term, $I(\nabla \cdot \mathbf{u})$, accounts for intensity variations induced by flow compression or expansion.

To adapt this motion prior to video generation, we must account for the distinct properties of visual data. Unlike compressible fluids where expansion necessitates density reduction to conserve mass, pixel intensity represents surface radiance, which remains stable under field-of-view changes. For example, during a camera zoom-in ($\nabla \cdot \mathbf{u} > 0$), object brightness does not decrease. Enforcing the full continuity equation would incorrectly penalize such natural geometric scaling unless the video artificially darkened. We therefore refine the motion constraint by omitting the divergence term, approximating the transport residual $\delta$ via the advective component alone:
\begin{align}
\label{eq:pixel_advection}
  \delta \approx \frac{\partial I}{\partial t} + \mathbf{u} \cdot \nabla I 
  = \frac{\partial I}{\partial t} + \frac{\partial h}{\partial t} \frac{\partial I}{\partial h} + \frac{\partial w}{\partial t} \frac{\partial I}{\partial w}.
\end{align}

While $\delta=0$ corresponds to the classical brightness constancy assumption\cite{horn1981determining}, we empirically observe that directly minimizing $\|\delta\|^2$ yields pathological, such as blurry and static motion. Real-world videos exhibit non-zero residuals due to complex lighting and sensor noise.
Nevertheless, $\delta$ serves as a critical motion-discriminative feature: its statistical distribution differs significantly between realistic motion and generated artifacts (e.g., high-frequency flickering). 

Instead of forcing $\delta \to 0$, we treat it as a dense feature map. 
As illustrated in Figure~\ref{fig:reward-model}(a), we compute the components of Eq.~\eqref{eq:pixel_advection} using differentiable operators. 
Spatial gradients $\frac{\partial I}{\partial h}$ and $\frac{\partial I}{\partial w}$ are extracted via Sobel operators, while the temporal derivative $\frac{\partial I}{\partial t}$ is approximated by finite differences between adjacent frames. The velocity field components, $u=\frac{\partial h}{\partial t}$ and $v=\frac{\partial w}{\partial t}$, are estimated using a pre-trained SEA-RAFT model~\cite{wang2024sea}. To account for estimation errors in the flow field, we concatenate the computed residual map $\delta$ with the flow confidence map provided by SEA-RAFT. This fused representation is processed by a Vision Transformer (ViT) trained to discriminate between the natural motion residuals of real videos and the artifacts of generated samples, thereby providing a robust, motion-discriminative signal for local temporal coherence.

\textbf{Long-term Motion Reward (LMR).}
Beyond local coherence, motion fidelity requires consistent global dynamics over extended horizons. As illustrated in Figure~\ref{fig:reward-model}(a), we evaluate long-term motion by tracking point trajectories $\tau_{\mathcal{P}} \in \mathbb{R}^{T \times M \times 2}$ initialized from a set of $M$ query points $\mathcal{P}$ sampled randomly from the first frame of the input video $\mathbf{X} \in \mathbb{R}^{T \times 3 \times H \times W}$. Unlike fixed grid sampling, this stochastic query initialization acts as an implicit ensemble over motion constraints, enhancing robustness against reward hacking.

We treat the evolution of pixel trajectories as a temporal stochastic process. To incorporate this motion-based inductive bias, we explicitly model the drift component via the instantaneous velocity of tracked points, $\tau_d[t] = \tau[t] - \tau[t-1]$. Trajectories are estimated using CoTracker3~\cite{karaev2025cotracker3}, which additionally provides visibility masks $\tau_v$ and confidence scores $\tau_c$. These components are concatenated to form a motion state descriptor $\mathbf{s} = [\tau_d, \tau_v, \tau_c]$. To further capture global structural dependencies, we augment this descriptor with dense correlation maps representing spatial affinities between query points and their temporal neighbors.

\textbf{Reward Training.}
Both the instantaneous and long-term reward models are parameterized as Vision Transformers (ViTs) and trained as binary discriminators to distinguish between real videos drawn from the dataset $p_{\text{data}}$ and synthetic samples generated by the diffusion model. Formally, we optimize the discriminator parameters $\omega$ by minimizing the binary cross-entropy loss:
\begin{equation}
    \label{eq:reward_loss}
    \mathcal{L}_{\text{rew}}(\omega) = -\mathbb{E}_{x \sim p_{\text{data}}} [\log D_\omega(x)] - \mathbb{E}_{\tilde{x} \sim p_\theta} [\log(1 - D_\omega(\tilde{x}))].
\end{equation}

The reward $r(\mathbf{x})$ is defined as the raw logit output of the discriminator, $r(\mathbf{x}) = \text{logit}(D_\omega(\mathbf{x}))$, clipped to a stable range $[r_{\min}, r_{\max}]$. We explicitly opt for raw logits over sigmoid-transformed probabilities to avoid gradient saturation, ensuring stronger learning signals particularly when the discriminator is confident (see ablation study).

\subsection{SHIFT: Smooth Hybrid Fine-tuning.}
\label{sec:shift_method}
To address aforementioned challenges in section \ref{sec: preliminary}, we present \textbf{SHIFT} (\textbf{S}mooth \textbf{H}ybr\textbf{I}d \textbf{F}ine-\textbf{T}uning), illustrated in Figure~\ref{fig:reward-model}(b), a data-regularized reinforcement learning framework designed for efficient video diffusion alignment. Instead of constraining the model to a drifting reference policy via Reverse KL, we anchor the model to the stationary ground-truth data distribution $p_{\text{data}}$ using \textit{Forward KL}. This is particularly valid for motion alignment, where we assume ground-truth video data possesses perfect motion fidelity.

\textbf{Deriving the Forward-KL Target.} Replacing the reference policy in Eq.~\eqref{eq:rl_optimal_solution} with stationary data distribution, we define optimal target $p^*(x)$ as:
\begin{equation}
    \label{eq:optimal_policy_target}
    p^*(x) \propto p_{\text{data}}(x) \exp(\frac{r(x)}{\beta}).
\end{equation}

While minimizing the Reverse KL between this target and $p_\theta$ is intractable due to the unnormalized density of $p^*$, minimizing the \textbf{Forward KL} divergence is straightforward:
\begin{equation}
    \label{eq:forward_kl_obj}
    \mathcal{J}(\theta) = D_{\text{KL}}(p^*(x) || p_\theta(x)) \equiv \arg \max_\theta \mathbb{E}_{x \sim p^*} [\log p_\theta(x)].
\end{equation}

\textbf{Optimization via Diffusion Loss Proxy.} Evaluating $\mathbb{E}_{x \sim p^*} [\dots]$ is non-trivial because we cannot directly sample from $p^*$. However, since $p^*$ is a product of the data distribution and the reward exponent, its probability mass concentrates in two regions: (1) the support of $p_{\text{data}}$, and (2) regions of high reward $r(x)$. We therefore approximate the gradient $\mathbb{E}_{x \sim p^*} [\nabla_\theta \log p_\theta(x)]$ with a mixture estimator over two accessible sources. An \emph{offline anchor} ($x \sim p_{\text{data}}$) samples directly from the dataset to preserve the generation quality and motion coherence of the base distribution, while an \emph{online exploration} term ($\tilde{x} \sim p_\theta$) uses the model's own rollouts, weighted by their relative advantage $A(\tilde{x})$, to cover high-reward regions favored by $\exp(r(\tilde{x})/\beta)$. This yields:
\begin{equation}
    \mathcal{J}(\theta) \approx \underbrace{\mathbb{E}_{\tilde{x} \sim p_\theta} [ \exp(\frac{A(\tilde{x})}{\beta}) \log p_\theta(\tilde{x}) ]}_{\text{Online Exploration}} + \underbrace{\mathbb{E}_{x \sim p_{\text{data}}} [\log p_\theta(x)]}_{\text{Offline Anchor}}.
\end{equation}

\begin{algorithm*}[t]
    \caption{SHIFT: Smooth Hybrid Fine-tuning}
    \label{alg:SHIFT}
    \textbf{Input:} pre-trained denoiser $\epsilon_{\text{base}}$, reward model $r_\omega$, 
    dataset $p_{\text{data}}$, iterations $N$, rollout size $B$, inner steps $K$, 
    temperature $\beta$. \\
    \textbf{Initialize:} $\theta \leftarrow \theta_{\text{base}}$.
    \begin{algorithmic}[1]
    \FOR{$n=1$ to $N$}
        \STATE Sample real data $\{x_0^b\}_{b=1}^B \sim p_{\text{data}}$ and fake samples $\{\tilde{x}_0^b\}_{b=1}^B \sim p_\theta$.
        \STATE Compute scaled rewards $\tilde{r}^b = r_\omega(\tilde{x}_0^b)/\beta$, $r^b = r_\omega(x_0^b)/\beta$, and mean $\bar r = \frac{1}{B}\sum_b \tilde{r}^b$.\STATE Compute advantages $\tilde{A}^b = \tilde{r}^b - \bar r$, $A^b = r^b - \bar r$.
        \FOR{$k=1$ to $K$}
            \STATE Update reward model by $\mathcal{L}_{\text{rew}} = -\mathbb{E}_b[\log D_\omega(x_0^b) + \log(1-D_\omega(\tilde{x}_0^b))]$.
            \STATE Sample $t\sim\mathcal{U}(0,T)$, $x_t\sim q(x_t|x_0)$, $\tilde{x}_t\sim q(x_t|\tilde{x}_0)$.
            \STATE Update generator with
            \[
            \mathcal{L}_{\text{SHIFT}}
            =
            \mathbb{E}_{b,t}\!\left[
            \tilde{A}^b\|\epsilon_\theta(\tilde{x}_t,t)-\tilde{\epsilon}\|^2
            + \|\epsilon_\theta(x_t,t)-\epsilon\|^2
            \right].
            \]
        \ENDFOR
    \ENDFOR
    \end{algorithmic}
    \end{algorithm*}

\textbf{Approximating Likelihood with Diffusion ELBO.}
We avoid the high memory cost of optimizing reverse trajectories by approximating the log-likelihood gradient $\nabla_\theta \log p_\theta(x)$ with the gradient of the Variational Lower Bound (ELBO). It is well-established that maximizing the log-likelihood is approximately equivalent to minimizing the weighted denoising error (please see the supplementary material for the detailed derivation):
\begin{equation}
    \nabla_\theta \log p_\theta(x) \approx - \nabla_\theta \mathcal{L}_{\text{diff}}(x) = - \nabla_\theta \mathbb{E}_{t, \epsilon} [ \|\epsilon_\theta(x_t, t) - \epsilon\|^2 ].
\end{equation}

Substituting this proxy into our hybrid objective yields the final loss function:
\begin{equation}
    \label{eq:shift_loss}
    \mathcal{L}_{\text{SHIFT}} = \mathbb{E}_{b,t} [ \underbrace{\tilde{A}^b \|\epsilon_\theta(\tilde{x}_t, t) - \tilde{\epsilon}\|^2}_{\text{Online RL Update}} + \underbrace{\|\epsilon_\theta(x_t, t) - \epsilon\|^2}_{\text{Offline Data Anchor}} ].
\end{equation}

Here, the first term reinforces self-generated trajectories ($\tilde{x}$) that achieve high rewards ($\tilde{A}^b > 0$), while the second term acts as a Supervised Fine-Tuning (SFT) regularizer preventing the policy from drifting out-of-distribution. This formulation provides a smooth integration of RL and SFT: when the advantage is zero, the algorithm reduces to standard fine-tuning; when rewards are high, it aggressively shifts the distribution toward desirable metrics. Detailed derivations of Eq. \eqref{eq:shift_loss} are provided in the supplementary material.

\subsection{Adversarial Advantages}
\label{sec:shift-gan}

\textbf{Advantages.} Similar to the GRPO \cite{shao2024deepseekmath}, at each iteration, we sample a minibatch of real videos $\{x_0^b\}_{b=1}^B\sim p_{\text{data}}$ and generate $\{\tilde{x}_0^b\}_{b=1}^B\sim p_\theta$ using a fixed discrete sampler. Then we compute temperature-scaled rewards $\tilde r^b=r_\omega(\tilde x_0^b)/\beta$ and $r^b=r_\omega(x_0^b)/\beta$, and center using the fake mean $\bar r=\frac1B\sum_b \tilde r^b$:
$\tilde A^b=\tilde r^b-\bar r$ and $A^b=r^b-\bar r$.
We use only recentering (no standard-deviation normalization) to avoid difficulty bias induced by small reward variance \cite{liu2025understanding}.

\textbf{Adversarial Online discriminator update.}
In unverifiable generation tasks, the generator distribution $p_\theta$ inevitably drifts from the initial training support, rendering static rewards unreliable and susceptible to hacking. To mitigate this, we employ an iterative adversarial update strategy. Periodically, we refresh the discriminator by keep optimizing Eq.~\eqref{eq:reward_loss}, using the most recent on-policy samples $\tilde{x}_0^b$ as negative examples against the ground-truth videos. In this framework, the reward approximates the log-density ratio $\log \frac{p_{\text{data}}(\mathbf{x})}{p_\theta(\mathbf{x})}$ in feature space, ensuring the signal remains accurate and robust to distribution shift throughout training. The complete SHIFT procedure is outlined in Algorithm \ref{alg:SHIFT}.

\textbf{Remark.}
While SHIFT's loss structure (Eq.~\eqref{eq:shift_loss}) bears a superficial resemblance to the reward-weighted regression plus SFT objective of Lee et al.~\cite{lee2023aligningtexttoimagemodelsusing}, the two methods differ in several fundamental respects. First, SHIFT derives its objective from a principled Forward KL formulation (Eq.~\eqref{eq:forward_kl_obj}), whereas Lee et al.\ adopt a heuristic combination of RWR and SFT without such theoretical grounding. Second, SHIFT replaces raw rewards with group-relative advantages and employs adversarial reward-model co-training to prevent reward hacking---mechanisms absent in Lee et al. Third, our reward models are fully automatic, motion-aware discriminators trained without any human annotation; this is essential for video motion alignment, where pixel-level temporal dynamics are practically infeasible for humans to label at scale.
Finally, although SHIFT alternates between generator and discriminator updates, it differs fundamentally from standard GAN training: the discriminator provides only a scalar reward signal used to compute advantages, and its gradient is never backpropagated into the generator. As shown by Pfau and Vinyals~\cite{pfau2016connecting}, GANs can be viewed as actor-critic methods where coupled gradient-based min-max optimization is the root cause of training instability; SHIFT sidesteps this by fully decoupling the two optimizations, retaining the discriminator's adaptive distribution-awareness without inheriting adversarial gradient dynamics.

\section{Experiments}

\textbf{Implementation.}
We validate SHIFT on two video diffusion models: SVD~\cite{blattmann2023stablevideodiffusionscaling} ($\sim$1.2B parameters), fine-tuned on DAVIS2017~\cite{pont20172017} (${\sim}$4K training clips after temporal segmentation), following the standard SVD fine-tuning protocol established by Track4Gen~\cite{jeong2025track4genteachingvideodiffusion}, and the Wan2.2 TI2V model ($\sim$5B parameters)~\cite{wan2025wan}, fine-tuned on WISA-80K~\cite{wang2025wisa}. For SVD, LoRA modules (rank 32) are applied only to temporal attention layers; for Wan2.2, LoRA (rank 32) is applied to all attention layers. Notably, unlike MDP-based baselines (e.g., FlowGRPO) that require stochastic samplers, SHIFT is sampler-agnostic. Full training and sampling configurations are provided in the supplementary material.

\textbf{Metrics.}
We adopt VBench-I2V~\cite{huang2024vbench}, grouping its sub-metrics into appearance-related (subject/background consistency, aesthetic/imaging quality, I2V subject/background) and motion-related (temporal flickering, motion smoothness, dynamic degree). We report category averages and the overall score on standard VBench-I2V, along with motion score~\cite{zhao2024identifying} and Fr\'{e}chet Video Distance (FVD).

\textbf{Reward model pre-training.}
Prior to RL fine-tuning, we pre-train the motion reward models as binary discriminators on paired real and generated videos. For each real video, we extract its first frame (optionally with the text prompt) and sample four synthetic videos from the base model as negatives. Details on data balancing and convergence monitoring are provided in the supplementary material.

\begin{table}[t]
  \centering
  \small
  \setlength{\tabcolsep}{3.5pt}
  \renewcommand\arraystretch{1.08}
    \caption{Quantitative results on VBench-I2V for the SVD model. FVD are tested on the DAVIS2017 validation set. Higher ($\uparrow$) is better; lower ($\downarrow$) is better.
  Best is \textbf{bold}; second-best is \uline{underlined}.}
  \resizebox{0.9\linewidth}{!}{%
  \begin{tabular}{lcccc|c|c}
    \toprule
    & \makecell[c]{VBench\\Appearance ($\uparrow$)}
    & \makecell[c]{VBench\\Motion ($\uparrow$)}
    & \makecell[c]{VBench\\Overall ($\uparrow$)}
    & \makecell[c]{Motion\\Score ($\uparrow$)}
    & \makecell[c]{FVD ($\downarrow$)}
    & \makecell[c]{Training\\Time ($\downarrow$)} \\
    \midrule
    Base            & 83.68 & 85.89 & 84.41 & 4.39 & 395.87 & -\\ 
    SFT             & \textbf{85.07} & 76.09 & 82.08 & 2.69 & \underline{322.81} & \textbf{1x} \\ 
    Track4Gen~\cite{jeong2025track4genteachingvideodiffusion}      & \underline{85.01} & 76.44 & 82.15 & 2.84 & \textbf{316.91} & \underline{1.07x} \\ 
    DenseDPO~\cite{wu2025densedpofinegrainedtemporalpreference}       & 83.73 & 86.03 & 84.50 & 4.34 & 399.73 & 2.40x \\ 
    FlowGRPO~\cite{liu2025flowgrpotrainingflowmatching} & 83.61 & \underline{86.67} & \underline{84.63} & \textbf{4.44} & 396.56 & 19.35x \\
    SHIFT (Ours)   & 83.69 & \textbf{86.70} & \textbf{84.69} & \underline{4.40} & 396.45 & 2.46x \\
    \bottomrule
  \end{tabular}%
  }
  \label{tab:svd_results}
\end{table}

\subsection{SVD fine-tuning}

For a fair comparison, all post-training methods are applied directly to the base model on the same hardware (64 GPUs), with per-epoch wall-clock time normalized to that of SFT ($1\times$); we set the buffer reuse factor $K{=}10$ for both SHIFT and FlowGRPO. Table~\ref{tab:svd_results} summarizes the results.

SFT and Track4Gen~\cite{jeong2025track4genteachingvideodiffusion} achieve the lowest FVD scores (322.81 and 316.91), but at a severe cost to motion quality: VBench Motion drops by 9.80 and 9.45 points, and Motion Score falls from 4.39 to 2.69 and 2.84 respectively, confirming the dynamic-degree degradation discussed in Section~\ref{sec:intro}. Per-dimension analysis shows that this is driven mostly by the collapse of the dynamic degree metric (from 0.67 to 0.33 for SFT), indicating near-static video generation. As noted by Ge et al.~\cite{ge2024content}, FVD is highly insensitive to temporal quality and biased toward per-frame appearance, rendering these seemingly strong FVD numbers misleading.

DenseDPO~\cite{wu2025densedpofinegrainedtemporalpreference} preserves motion quality (VBench Motion 86.03, Motion Score 4.34) but slightly worsens FVD (399.73 vs.\ 395.87) and yields only marginal overall gains (+0.09). FlowGRPO~\cite{liu2025flowgrpotrainingflowmatching} further improves motion, achieving the highest Motion Score (4.44) and strong gains in VBench Motion (86.67, +0.78) and Overall (84.63, +0.22); however, it slightly degrades appearance (83.61 vs.\ 83.68) and is considerably more expensive ($19.35\times$ vs.\ $2.46\times$). In contrast, SHIFT achieves the best VBench Motion (86.70, +0.03 over FlowGRPO) and Overall (84.69, +0.06 over FlowGRPO) while maintaining appearance on par with the base model (83.69 vs.\ 83.68) and comparable FVD (396.45). SHIFT is the only method that simultaneously improves motion fidelity, overall quality, and motion magnitude without degrading any individual metric, at a fraction of FlowGRPO's computational budget.

Regarding training efficiency, Track4Gen adds a lightweight correspondence loss atop SFT ($1.07\times$). DenseDPO ($2.40\times$) and SHIFT ($2.46\times$) incur comparable costs from winner--loser pair processing and adversarial reward-model updates, respectively; both operate entirely in the forward diffusion process without optimizing through the reverse denoising trajectory. FlowGRPO is substantially more expensive ($19.35\times$, ${\approx}7.9\times$ that of SHIFT) because it formulates denoising as a Markov decision process and recomputes per-timestep log-probabilities across all 25 sampling steps, requiring much more UNet forward passes per step.

\begin{figure}[t]
  \centering
  \begin{minipage}[t]{0.48\linewidth}
    \centering
    \includegraphics[width=\linewidth]{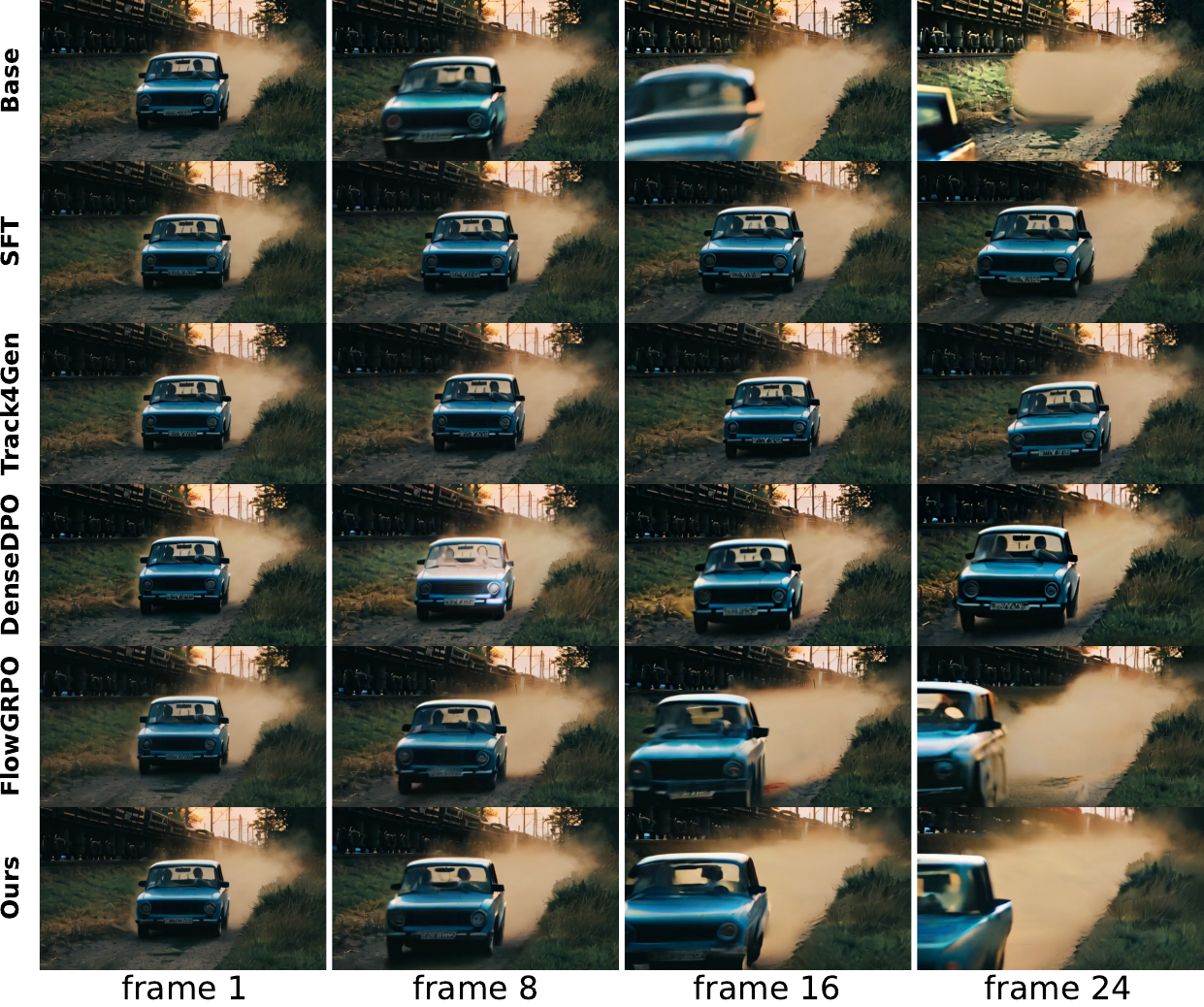}
  \end{minipage}
  \hfill
  \begin{minipage}[t]{0.48\linewidth}
    \centering
    \includegraphics[width=\linewidth]{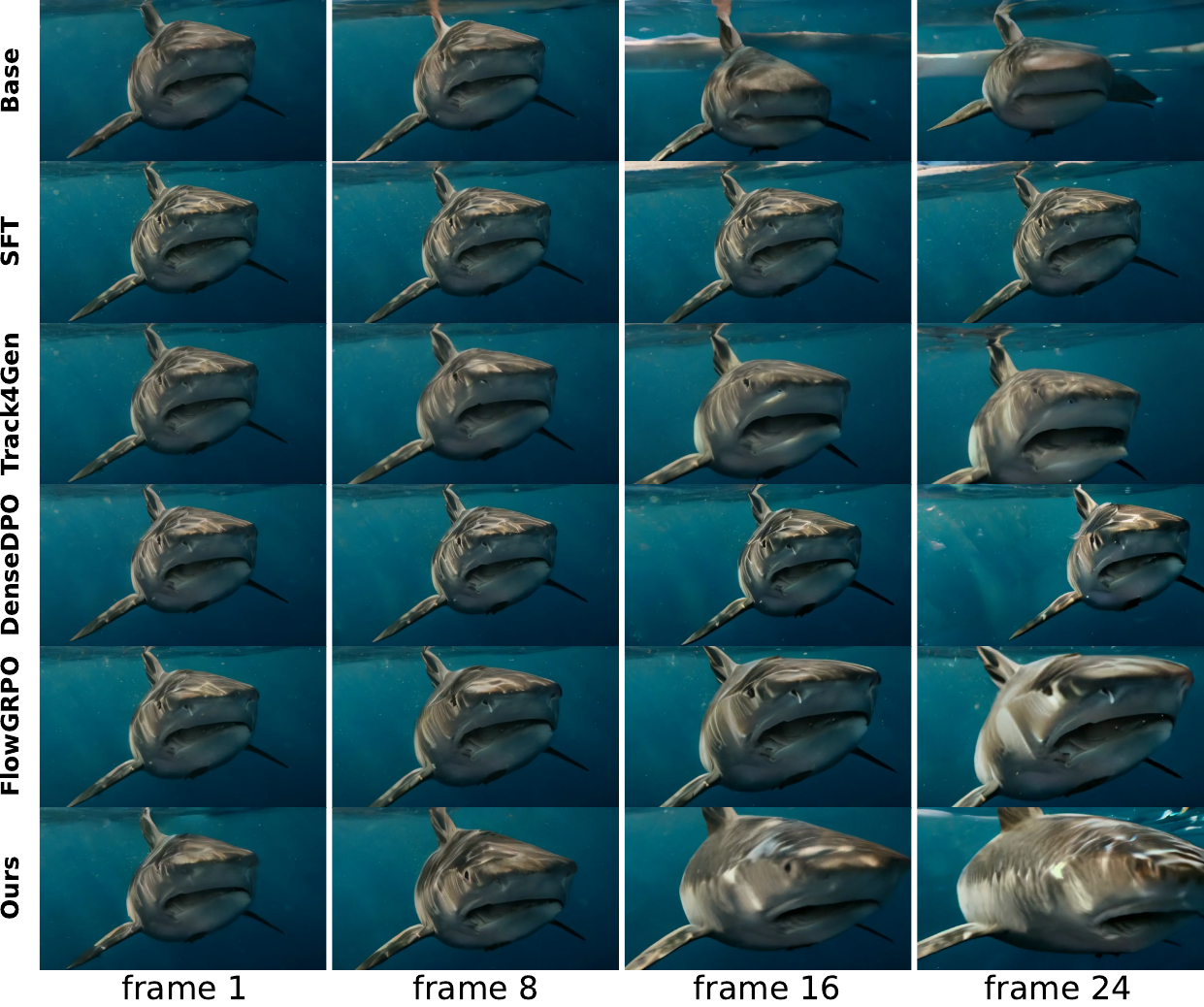}
  \end{minipage}
  \caption{Qualitative comparison of VBench-I2V Standard test examples generated by the base SVD model and different fine-tuned SVD variants.}
  \label{fig:qualitative_comparison}
\end{figure}

Please see Figure~\ref{fig:qualitative_comparison} for qualitative examples showing that SHIFT improves motion dynamics while preserving strong visual quality. More comparisons are provided in the supplementary material.

\subsection{Wan2.2-TI2V fine-tuning}

\begin{figure}[t]
  \centering
  \begin{minipage}[t]{0.48\linewidth}
    \centering
    \includegraphics[width=\linewidth]{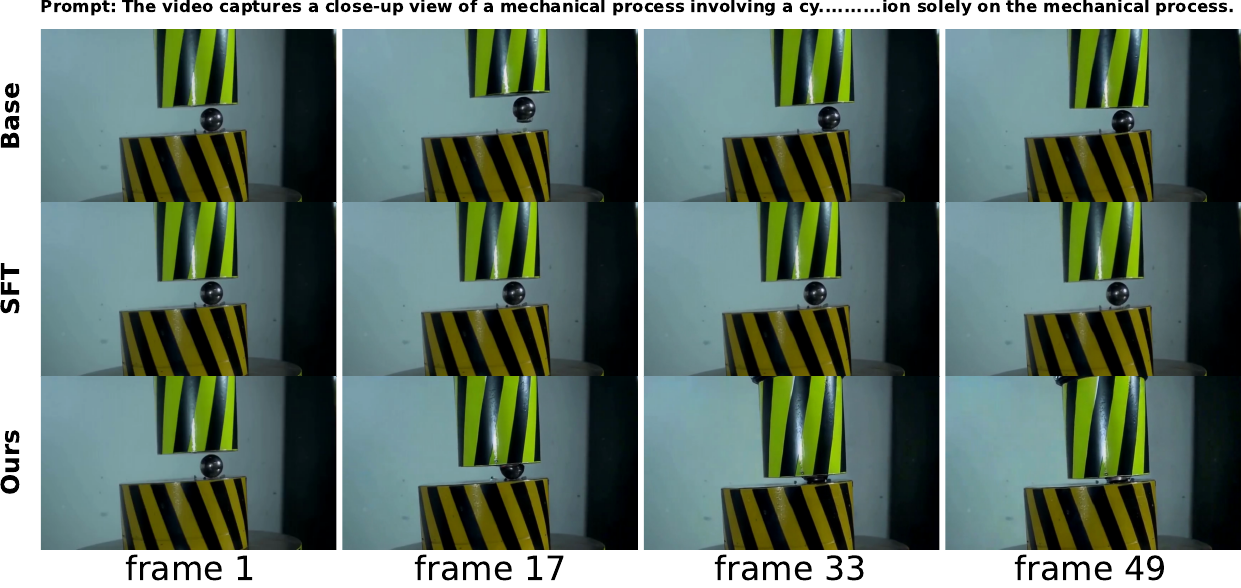}
  \end{minipage}
  \hfill
  \begin{minipage}[t]{0.48\linewidth}
    \centering
    \includegraphics[width=\linewidth]{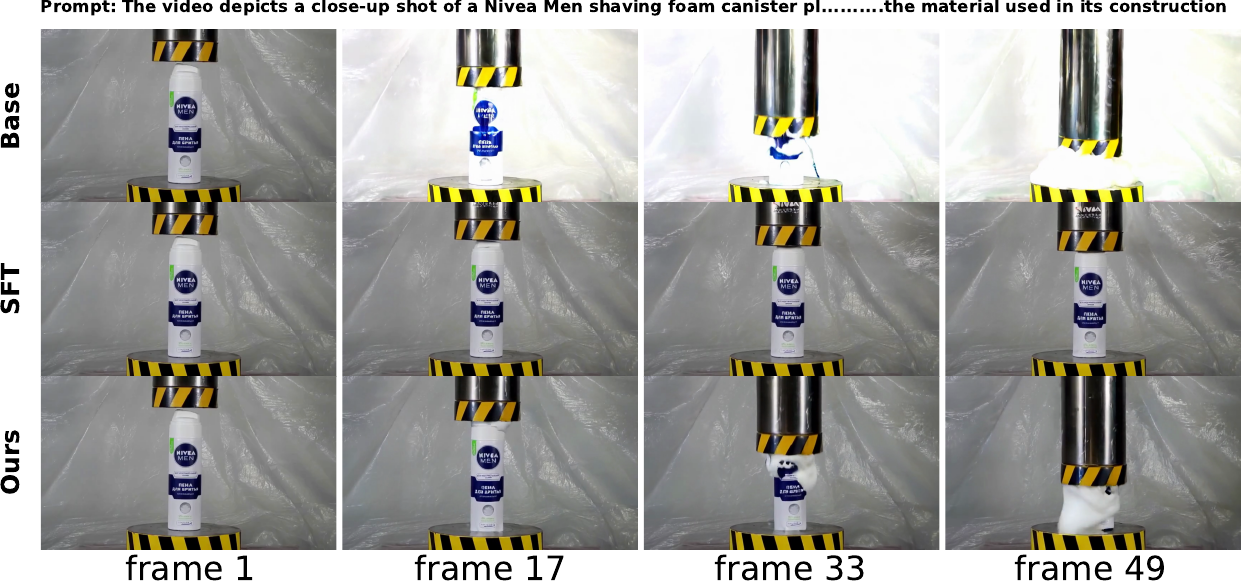}
  \end{minipage}
  \caption{Qualitative comparison on the WISA-80K validation set for the base Wan2.2-TI2V model and fine-tuned variants. SHIFT generates more realistic motion.}
  \label{fig:wan_qualitative_comparison}
\end{figure}

We further evaluate SHIFT on the 5B Wan2.2-TI2V model fine-tuned on the deformation subset of WISA-80K, which contains relatively rare real-world physical phenomena. Table~\ref{tab:wan_results_standard} shows a trend consistent with the SVD experiments: SFT improves appearance and FVD but substantially degrades motion (Motion Score 1.94 vs.\ 2.96). In contrast, SHIFT improves the overall VBench score (86.40 vs.\ 86.27) with modest gains in both appearance and motion while keeping FVD comparable to the base model. Figure~\ref{fig:wan_qualitative_comparison} provides qualitative examples, and additional results are included in the supplementary material.

\begin{table}[t]
  \centering
  \scriptsize
  \setlength{\tabcolsep}{3.5pt}
  \renewcommand\arraystretch{1.08}
  \caption{Quantitative results on VBench-I2V Standard for the Wan2.2 TI2V model. Higher ($\uparrow$) is better; lower ($\downarrow$) is better.
  Best is \textbf{bold}; second-best is \uline{underlined}.}
  \resizebox{0.9\linewidth}{!}{
  \begin{tabular}{lcccc|c}
    \toprule
    & \makecell[c]{VBench\\Appearance ($\uparrow$)}
    & \makecell[c]{VBench\\Motion ($\uparrow$)}
    & \makecell[c]{VBench\\Overall ($\uparrow$)}
    & \makecell[c]{Motion\\Score ($\uparrow$)}
    & \makecell[c]{FVD ($\downarrow$)} \\
    \midrule
    Base            & 86.29 & \underline{86.25} & \underline{86.27} & \textbf{2.96} & 284.68  \\ 
    SFT             & \textbf{87.14} & 83.34 & 85.87 & 1.94 & \textbf{151.44} \\ 
    SHIFT (ours)   & \underline{86.38} & \textbf{86.45} & \textbf{86.40} & \underline{2.93} & \underline{282.24}  \\
    \bottomrule
  \end{tabular}
  }
  \label{tab:wan_results_standard}
\end{table}

\subsection{Ablation study}

\begin{table}[t]
  \centering
  \small
  \setlength{\tabcolsep}{3.5pt}
  \renewcommand\arraystretch{1.08}
  \caption{Ablation of SHIFT components evaluated at the same epoch 7.
  Higher ($\uparrow$) is better; lower ($\downarrow$) is better.
  Best is \textbf{bold}; second-best is \uline{underlined}.}
  \resizebox{0.9\linewidth}{!}{%
  \begin{tabular}{l cccc|cccc|c}
    \toprule
    & \makecell[c]{IMR}
    & \makecell[c]{Noise\\Align}
    & \makecell[c]{Adv.\\RM}
    & \makecell[c]{LMR}
    & \makecell[c]{VBench\\Appearance\ ($\uparrow$)}
    & \makecell[c]{VBench\\Motion ($\uparrow$)}
    & \makecell[c]{VBench\\Overall ($\uparrow$)}
    & \makecell[c]{Motion\\Score ($\uparrow$)}
    & \makecell[c]{FVD\\($\downarrow$)} \\
    \midrule
    A0 & \checkmark &  &  &  & 81.62 & \textbf{91.01} & \uline{84.75} & \underline{4.29} & 502.60 \\
    A1 & \checkmark & \checkmark &  &  & 82.53 & 87.18 & 84.08 & 3.75 & 472.20 \\
    A2 & \checkmark & \checkmark & \checkmark &  & \underline{83.06} & 87.28 & 84.47 & 3.75 & \underline{451.20} \\
    A3 & \checkmark & \checkmark &  & \checkmark & 82.48 & \uline{89.31} & \textbf{84.76} & 4.14 & 460.00 \\
    A4 & \checkmark & \checkmark & \checkmark & \checkmark & \textbf{83.63} & 86.98 & \underline{84.75} & \textbf{4.35} & \textbf{401.76} \\
    \bottomrule
  \end{tabular}%
  }
  \label{tab:shift_ablation}
\end{table}

\begin{figure}[t]
  \centering
  \includegraphics[width=\linewidth]{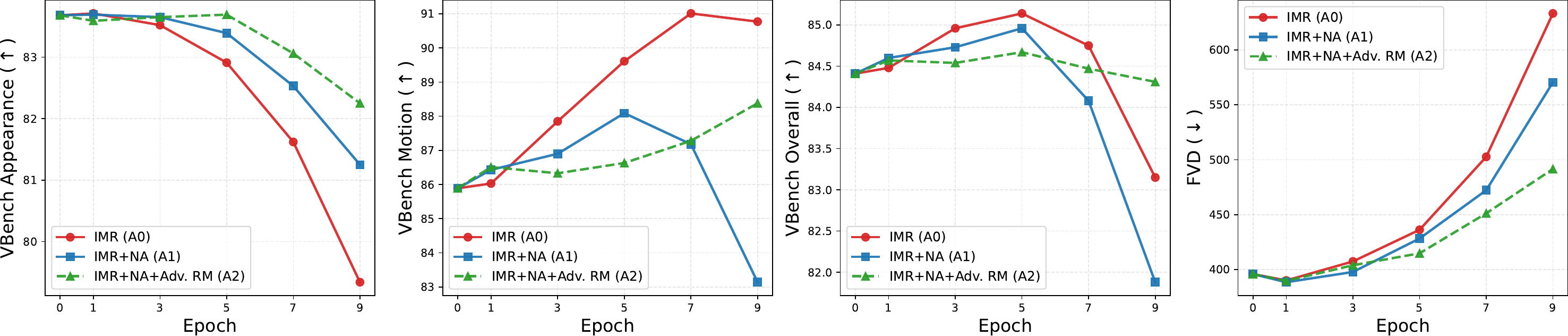}
  \caption{Ablation of SHIFT components over training epochs. All models start from the same pre-trained SVD checkpoint. Adding Noise Alignment (NA) strengthens the SFT anchor regularization, while the Adversarial RM further stabilizes training to avoid reward hacking.}
  \label{fig:imr_ablation}
\end{figure}

We analyze the contribution of each SHIFT component by progressively adding regularization mechanisms to the Instantaneous Motion Reward baseline (A0 in Table~\ref{tab:shift_ablation}): first Noise Alignment (A1), then Adversarial Reward Model updates (A2). These two mechanisms serve primarily to stabilize training and prevent reward hacking rather than to boost motion quality directly. Since the Long-term Motion Reward provides an orthogonal motion supervision signal, we evaluate it separately by adding LMR to the A1 baseline (A3). The full model (A4) combines all four components. Figure~\ref{fig:imr_ablation} summarizes the training dynamics.

\textbf{IMR baseline (A0).}
Training with IMR alone yields the highest VBench Motion among all variants (91.01) but at a steep cost: Appearance falls to 81.62 and FVD rises to 502.60, indicating reward hacking. As shown in Figure~\ref{fig:imr_ablation}, this degradation worsens over continued training. We attribute this to a \emph{noise-level mismatch} between the SFT anchor and the advantage-weighted regression (AWR) term: higher noise levels predominantly govern coarse structure and motion, while lower levels control fine visual details~\cite{choi2022perception}. When the two loss terms sample noise levels independently, AWR aggressively optimizes motion at high noise scales while SFT regularization is diluted across random scales, leaving appearance unprotected.

\textbf{+\,Noise Alignment (A1).}
To address this mismatch, Noise Alignment (NA) couples the noise schedules of SFT and AWR by sharing the same sampled noise level within each training step. As shown in Table~\ref{tab:shift_ablation}, NA improves Appearance to 82.53 (+0.91 over A0) and reduces FVD to 472.20 (vs.\ 502.60). Figure~\ref{fig:imr_ablation}(a,\,d) further confirms that NA produces more stable training dynamics with substantially less reward hacking, confirming that aligning noise strengthens the SFT anchor at scales where the AWR objective exerts strong gradient.

\textbf{+\,Adversarial Reward Model (A2).}
While NA mitigates the noise-level mismatch, the static reward model becomes increasingly exploitable as the generator distribution shifts. Adding adversarial reward-model updates, which periodically optimizes the discriminator on the latest generated samples and real samples, further improves results: Appearance rises to 83.06, FVD decreases to 451.20, and VBench Overall reaches 84.47, outperforms A1. Figure~\ref{fig:imr_ablation} confirms that A2 exhibits the most stable training trajectory, demonstrating that adversarial co-training enables sustained improvement by mitigating reward hacking.

\textbf{+\,Long-term Motion Reward (A3).}
Adding LMR to the A1 baseline recovers the motion quality that NA alone sacrifices for stability. A3 improves VBench Motion by +2.13 and Motion Score by +0.39 over A1, while preserving comparable appearance (82.48 vs.\ 82.53). Notably, the dynamic degree---the VBench dimension most indicative of meaningful object motion---rises from 68.54 to 75.93, confirming that LMR's trajectory-level supervision prevents the model from satisfying the instantaneous reward through superficial artifacts (e.g., texture flickering) rather than genuine dynamics. A3 also achieves the best VBench Overall (84.76), demonstrating that IMR and LMR provide complementary signals: IMR enforces local temporal coherence via optical-flow consistency, while LMR anchors global trajectory plausibility over the full horizon.

\textbf{Full model (A4).}
Combining all four components yields the full SHIFT model, which largely prevents reward hacking and achieves the best overall balance between appearance and motion.
Compared to A1, A4 recovers +1.10 in Appearance (83.63 vs.\ 82.53) and +0.60 in Motion Score (4.35 vs.\ 3.75), while reducing FVD by 70.44.
Compared to A3, A4 further improves Appearance by +1.15 and reduces FVD by 58.24, confirming that adversarial reward training is essential for preventing distributional drift even when LMR is present.
A4 attains the best Appearance, Motion Score, and FVD among all variants, trailing A3 in VBench Overall by only 0.01.
These results reveal a clear division of roles: IMR and LMR provide the motion supervision signal at complementary temporal scales, while NA and Adv.\ RM serve as regularization mechanisms that prevent the optimization from degenerating into reward hacking. Neither NA nor Adv.\ RM improves motion quality per se; instead, they ensure that the motion gains from IMR and LMR translate into genuine quality improvements rather than distributional drift.

\textbf{Logit vs.\ Sigmoid Reward.}
Our reward models are binary discriminators producing unbounded logits. We compare using the raw logit $r(\mathbf{x}) = f_\omega(\mathbf{x})$ versus the sigmoid probability $r(\mathbf{x}) = \sigma(f_\omega(\mathbf{x}))$ as the reward signal. The sigmoid saturates near 0 or 1 for large-magnitude logits, compressing the group-relative advantages $\tilde{A}^b = \tilde{r}^b - \bar{r}$ to near-zero variance and effectively stalling LoRA optimization. Raw logits preserve the full dynamic range of the discriminator's confidence, yielding well-separated advantages and meaningful gradient updates. Since raw logits are unbounded, we clip the computed advantages to a fixed range for stability, analogous to the clipping in PPO~\cite{schulman2017proximalpolicyoptimizationalgorithms} and GRPO~\cite{shao2024deepseekmath}. We adopt clipped raw-logit rewards throughout all experiments.

Additional ablations on temperature $\beta$ and replay factor $K$ are provided in the supplementary material. In summary, smaller $\beta$ accelerates reward optimization but increases reward hacking, while moderate $\beta$ offers a better stability--quality tradeoff. Reducing $K$ from 10 to 5 causes minor metric changes but increases wall-clock cost, indicating that SHIFT is robust to moderate off-policy replay. We also ablate the reward model architecture (ViT vs.\ ResNet) and the LMR query point sampling strategy (uniform grid vs.\ random sampling) in the supplementary material.
\section{Conclusion}
We address dynamic-degree degradation in image-conditioned video diffusion models by introducing pixel-motion rewards derived from optical-flow and point-track features, capturing both instantaneous temporal coherence and long-term global motion dynamics. Built on these rewards, SHIFT unifies supervised anchoring with advantage-weighted updates in a sampler-agnostic framework, and incorporates adversarial reward-model co-training and noise-level alignment to mitigate reward hacking. Experiments show that SHIFT improves motion fidelity without sacrificing appearance quality.

\noindent\textbf{Acknowledgements.} This work was supported in part by the NSFC under Grant 92570001; in part by the NSFC Joint Fund and Special Fund under Grants U25B6003 and 62550004. This work was supported by Ant Group Research Fund. 

\bibliographystyle{splncs04}
\bibliography{reference}

\appendix
\renewcommand{\thetable}{A\arabic{table}}
\renewcommand{\thefigure}{A\arabic{figure}}
\renewcommand{\theHtable}{appendix.\arabic{table}}
\renewcommand{\theHfigure}{appendix.\arabic{figure}}
\setcounter{table}{0}
\setcounter{figure}{0}
\section{Supplementary Material}

\subsection{Preliminaries: Diffusion Models}
\label{sec:app_prelim}

The Denoising Diffusion Probabilistic Model (DDPM) framework~\cite{hoddpm} defines a fixed forward process that gradually corrupts data with Gaussian noise, and a learned reverse process that iteratively denoises to generate samples.

\textbf{Forward Process.}
Given a data sample $\mathbf{x}_0 \sim p_{\text{data}}(\mathbf{x})$, the forward noising process over $T$ timesteps is a Markov chain:
\begin{equation}
q(\mathbf{x}_{1:T} | \mathbf{x}_0) := \prod_{t=1}^{T} q(\mathbf{x}_t | \mathbf{x}_{t-1}), \quad q(\mathbf{x}_t | \mathbf{x}_{t-1}) = \mathcal{N}(\mathbf{x}_t;\, \sqrt{1-\beta_t}\,\mathbf{x}_{t-1},\, \beta_t \mathbf{I}),
\end{equation}
where $\{\beta_t \in (0,1)\}_{t=1}^{T}$ is a fixed variance schedule.
Defining $\alpha_t := 1 - \beta_t$ and $\bar{\alpha}_t := \prod_{s=1}^{t} \alpha_s$, we can sample $\mathbf{x}_t$ at arbitrary $t$ in closed form:
\begin{equation}
\label{eq:app_forward}
\mathbf{x}_t = \sqrt{\bar{\alpha}_t}\,\mathbf{x}_0 + \sqrt{1 - \bar{\alpha}_t}\,\boldsymbol{\epsilon}, \quad \boldsymbol{\epsilon} \sim \mathcal{N}(\mathbf{0}, \mathbf{I}).
\end{equation}

\textbf{Reverse Process.}
The generative model learns a parameterized reverse chain starting from $p(\mathbf{x}_T) = \mathcal{N}(\mathbf{0}, \mathbf{I})$:
\begin{equation}
p_\theta(\mathbf{x}_{0:T}) := p(\mathbf{x}_T) \prod_{t=1}^{T} p_\theta(\mathbf{x}_{t-1} | \mathbf{x}_t), \quad p_\theta(\mathbf{x}_{t-1} | \mathbf{x}_t) = \mathcal{N}\!\left(\mathbf{x}_{t-1};\, \boldsymbol{\mu}_\theta(\mathbf{x}_t, t),\, \boldsymbol{\Sigma}_\theta(\mathbf{x}_t, t)\right).
\end{equation}
A practical parameterization trains a network $\boldsymbol{\epsilon}_\theta(\mathbf{x}_t, t)$ to predict the noise $\boldsymbol{\epsilon}$, from which the mean is derived as:
\begin{equation}
\boldsymbol{\mu}_\theta(\mathbf{x}_t, t) = \frac{1}{\sqrt{\alpha_t}} \left( \mathbf{x}_t - \frac{\beta_t}{\sqrt{1 - \bar{\alpha}_t}}\,\boldsymbol{\epsilon}_\theta(\mathbf{x}_t, t) \right).
\end{equation}
The variance can be fixed to $\tilde{\beta}_t = \frac{1 - \bar{\alpha}_{t-1}}{1 - \bar{\alpha}_t}\,\beta_t$ or learned~\cite{nichol2021improveddenoisingdiffusionprobabilistic}.

\textbf{Training Objective.}
The simplified training objective is the mean-squared error between the predicted and true noise:
\begin{equation}
\label{eq:app_simple_loss}
\mathcal{L}_{\text{simple}}(\theta) = \mathbb{E}_{\mathbf{x}_0 \sim p_{\text{data}},\, t \sim \mathcal{U}(1,T),\, \boldsymbol{\epsilon} \sim \mathcal{N}(\mathbf{0},\mathbf{I})} \left[ \| \boldsymbol{\epsilon} - \boldsymbol{\epsilon}_\theta(\mathbf{x}_t, t) \|^2 \right],
\end{equation}
which is equivalent to denoising score matching~\cite{song2020generativemodelingestimatinggradients}.

\textbf{Sampling.}
Samples are generated by iteratively denoising from $\mathbf{x}_T \sim \mathcal{N}(\mathbf{0}, \mathbf{I})$:
\begin{equation}
\mathbf{x}_{t-1} = \frac{1}{\sqrt{\alpha_t}} \left( \mathbf{x}_t - \frac{\beta_t}{\sqrt{1 - \bar{\alpha}_t}}\,\boldsymbol{\epsilon}_\theta(\mathbf{x}_t, t) \right) + \sigma_t \mathbf{z}, \quad \mathbf{z} \sim \mathcal{N}(\mathbf{0}, \mathbf{I}),
\end{equation}
for $t = T, \ldots, 1$, where $\sigma_t = \sqrt{\tilde{\beta}_t}$ for the stochastic sampler.
DDIM~\cite{ddim} provides an alternative deterministic sampler that accelerates generation.

\subsection{Reinforcement Learning for Diffusion Fine-Tuning}
\label{sec:app_rl}

Applying RL to align video diffusion models with non-differentiable objectives introduces challenges beyond those in image generation: storing full denoising trajectories incurs prohibitive memory costs, methods are often tightly coupled to specific samplers, and long temporal horizons exacerbate credit assignment~\cite{domingoenrich2025adjointmatchingfinetuningflow,liu2025timerewarderlearningdensereward,wu2025densedpofinegrainedtemporalpreference}.
In unverifiable environments the risk of reward hacking---where generators exploit imperfections in reward models---is especially pronounced~\cite{wu2025rewarddancerewardscalingvisual,amodei2016concreteproblemsaisafety}.
Below we review the two main paradigms and existing mitigation strategies.

\subsubsection{MDP-Based Approaches.}
\label{subsec:app_mdp}

DDPO~\cite{black2024trainingdiffusionmodelsreinforcement} and DPOK~\cite{fan2023dpokreinforcementlearningfinetuning} formulate the reverse denoising process as a multi-step MDP.
Given a conditioning signal $c$, the denoising trajectory $\tau = (\mathbf{x}_T, \ldots, \mathbf{x}_0)$ defines a policy $\pi_\theta(\mathbf{x}_{t-1} | \mathbf{x}_t, c) = p_\theta(\mathbf{x}_{t-1} | \mathbf{x}_t, c)$ with sparse reward $r(\mathbf{x}_0, c)$ at the terminal state.
The RL objective is:
\begin{equation}
J(\theta) = \mathbb{E}_{c \sim p(c),\, \tau \sim p_\theta(\tau|c)} \left[ r(\mathbf{x}_0, c) \right].
\end{equation}
Applying the REINFORCE estimator~\cite{Williams2004SimpleSG} yields:
\begin{equation}
\label{eq:app_reinforce}
\nabla_\theta J(\theta) = \mathbb{E}_{\tau \sim p_\theta(\tau|c)} \left[ \left( \sum_{t=1}^{T} \nabla_\theta \log \pi_\theta(\mathbf{x}_{t-1} | \mathbf{x}_t, c) \right) r(\mathbf{x}_0, c) \right].
\end{equation}
To reduce variance and enable multiple gradient steps per rollout, importance-weighted variants with PPO-style clipping are used~\cite{schulman2017proximalpolicyoptimizationalgorithms}.
The key limitation is computational: generating and storing complete reverse trajectories is expensive for high-resolution video, and the training procedure is locked to a specific SDE sampler.

\subsubsection{Forward-Process Methods.}
\label{subsec:app_forward}

An alternative paradigm avoids sequential RL altogether by operating on the forward diffusion process, decoupling reward optimization from the Markovian structure of denoising.

\textbf{Reward-Weighted Regression (RWR).}
RWR~\cite{peters2007reinforcement,lee2023aligningtexttoimagemodelsusing} treats generation as a single-step decision and minimizes a weighted negative log-likelihood:
\begin{equation}
\label{eq:app_rwr}
\mathcal{L}_{\text{RWR}}(\theta) = -\mathbb{E}_{c,\, \mathbf{x}_0 \sim \pi_{\text{old}}} \left[ w(r(\mathbf{x}_0, c)) \cdot \log \pi_\theta(\mathbf{x}_0|c) \right],
\end{equation}
where $w(\cdot)$ is a non-negative weighting function (e.g., $w(r) \propto \exp(\beta r)$).
Since the marginal likelihood $\log \pi_\theta(\mathbf{x}_0|c)$ is intractable, the standard denoising loss is substituted, but this lacks theoretical justification for maximizing $r(\mathbf{x}_0, c)$~\cite{black2024trainingdiffusionmodelsreinforcement}.
Lee et al.~\cite{lee2023aligningtexttoimagemodelsusing} combine RWR with SFT for image diffusion, but rely on a purely offline dataset and raw rewards; SHIFT instead uses online samples with adversarial GRPO-style advantages.

\textbf{Adjoint Matching.}
Adjoint Matching~\cite{domingoenrich2025adjointmatchingfinetuningflow} provides a continuous-time perspective for flow-based models. The optimal fine-tuned velocity field satisfies:
\begin{equation}
v^*(\mathbf{x}_t, t) = v(\mathbf{x}_t, t) + (1 - t)\, \nabla_{\mathbf{x}_t} \log \mathbb{E}_{p(\mathbf{x}_0|\mathbf{x}_t)}[\exp(\beta\, r(\mathbf{x}_0))],
\end{equation}
which is approximated via a reweighted flow matching loss $\mathcal{L}_{\text{AM}}(\theta) = \mathbb{E}_{t, \mathbf{x}_0, \boldsymbol{\epsilon}} [ w(\mathbf{x}_0, r) \cdot \| v_\theta(\mathbf{x}_t, t) - (\dot{\alpha}_t \mathbf{x}_0 + \dot{\sigma}_t \boldsymbol{\epsilon}) \|^2 ]$.

\textbf{DiffusionNFT.}
DiffusionNFT~\cite{zheng2025diffusionnft} interprets the reward as an optimality probability and defines implicit positive/negative policies $\pi_{+} \propto r \cdot \pi_{\text{old}}$ and $\pi_{-} \propto (1-r) \cdot \pi_{\text{old}}$.
A contrastive loss pulls the model toward $\pi_{+}$ and away from $\pi_{-}$:
\begin{equation}
\mathcal{L}_{\text{NFT}}(\theta) = \mathbb{E}_{t, \mathbf{x}_0, \boldsymbol{\epsilon}} \left[ r\, \| v_\theta^{+} - \mathbf{v} \|^2 + (1 - r)\, \| v_\theta^{-} - \mathbf{v} \|^2 \right],
\end{equation}
where $v_\theta^{+} = (1-\beta)v_{\text{old}} + \beta v_\theta$, $v_\theta^{-} = (1+\beta)v_{\text{old}} - \beta v_\theta$, and $\mathbf{v}$ is the flow matching target.

Compared to MDP-based methods, forward-process approaches offer better training stability, lower memory requirements (only $\mathbf{x}_0$ and rewards are stored), and solver agnosticism; however, the ELBO substitution introduces approximation error.

\subsubsection{Reward Hacking Mitigation.}
\label{subsec:app_hacking}

The risk of reward hacking is especially severe in video generation due to its high dimensionality and sparse supervision~\cite{wu2025rewarddancerewardscalingvisual,amodei2016concreteproblemsaisafety}.
Existing mitigations include:
(i)~KL penalties or trust region methods that constrain policy updates~\cite{jaques2019wayoffpolicybatchdeep};
(ii)~GRPO~\cite{shao2024deepseekmath} and batch-wise relative advantages that reduce sensitivity to absolute reward scaling~\cite{liu2025flowgrpotrainingflowmatching};
(iii)~iterative preference optimization that alternates between policy and reward model updates;
and (iv)~adversarial frameworks such as RL Tango~\cite{zha2025rl} that incorporate discriminator-based rewards evolving alongside the generator~\cite{ma2024videovideogenerativeadversarial,lin2025autoregressiveadversarialposttrainingrealtime}.
SHIFT integrates forward-process fine-tuning with adversarial reward model updates, combining the stability of SFT-based regularization with the adaptability of adversarial co-training.

\subsection{Derivation of the SHIFT Objective}
\label{sec:app_derivation}

We provide a detailed derivation connecting the constrained RL formulation (Eq.~1 in the main paper) to the final SHIFT loss (Eq.~12 of the main paper).

\subsubsection{Constrained RL and Optimal Policy.}

The standard objective is to maximize the expected reward subject to a trust-region constraint:
\begin{equation}
\max_\theta\; \mathbb{E}_{x_{0:T} \sim p_\theta}[r(x_0)] \quad \text{s.t.} \quad D_{\mathrm{KL}}(p_\theta \| p_{\mathrm{ref}}) \le \gamma.
\end{equation}
Introducing the Lagrange multiplier $\beta > 0$ and enforcing the KKT conditions~\cite{nair2021awacacceleratingonlinereinforcement, peters2007reinforcement}, the unconstrained Lagrangian becomes:
\begin{equation}
\label{eq:app_lagrangian}
\mathcal{L} \propto -\mathbb{E}_{x_{0:T} \sim p_\theta}[r(x_0)] + \beta\, D_{\mathrm{KL}}(p_\theta \| p_{\mathrm{ref}}).
\end{equation}
Setting the functional derivative $\delta \mathcal{L} / \delta p_\theta = 0$ yields the optimal policy in closed form:
\begin{equation}
\label{eq:app_optimal}
p_\theta^* \propto p_{\mathrm{ref}} \exp\!\left(\frac{r(x_0)}{\beta}\right).
\end{equation}
The corresponding policy gradient is:
\begin{align}
\label{eq:app_pg}
\nabla_\theta \mathcal{L}
&= -\mathbb{E}_{x_{0:T}}\!\left[\nabla_\theta \log p_\theta(x_{0:T}) \cdot r(x_0)\right] + \nabla_\theta D_{\mathrm{KL}}(p_\theta \| p_{\mathrm{ref}}) \nonumber \\
&= -\mathbb{E}_{x_{0:T}}\!\left[\sum_{t=1}^{T} \nabla_\theta \log p_\theta(x_{t-1}|x_t) \cdot r(x_0)\right] + \nabla_\theta D_{\mathrm{KL}}(p_\theta \| p_{\mathrm{ref}}).
\end{align}

\subsubsection{From Reverse KL to Forward KL with Data Anchor.}

Standard methods (DDPO, FlowGRPO) minimize the Reverse KL $D_{\mathrm{KL}}(p_\theta \| p^*)$, which requires on-policy sampling and evaluation of the intractable ratio $p^*/p_\theta$.
SHIFT replaces the drifting reference policy $p_{\mathrm{ref}}$ with the stationary data distribution $p_{\mathrm{data}}$, defining the optimal target as:
\begin{equation}
p^*(x) \propto p_{\mathrm{data}}(x) \exp\!\left(\frac{r(x)}{\beta}\right).
\end{equation}
Instead of the intractable Reverse KL, we minimize the \textbf{Forward KL}:
\begin{equation}
\label{eq:app_fkl}
\mathcal{J}(\theta) = D_{\mathrm{KL}}(p^* \| p_\theta) = \mathbb{E}_{x \sim p^*}\!\left[-\log p_\theta(x)\right] + \mathrm{const}.
\end{equation}
Minimizing $\mathcal{J}$ is equivalent to maximizing $\mathbb{E}_{x \sim p^*}[\log p_\theta(x)]$.

\subsubsection{Mixture Estimator.}

We cannot sample from $p^*$ directly.
Substituting $p^*(x) \propto p_{\mathrm{data}}(x) \exp(r(x)/\beta)$, the expectation becomes:
\begin{equation}
\mathbb{E}_{x \sim p^*}[\log p_\theta(x)] = \frac{1}{Z} \int p_{\mathrm{data}}(x)\, \exp\!\left(\frac{r(x)}{\beta}\right) \log p_\theta(x)\, dx,
\end{equation}
where $Z = \int p_{\mathrm{data}}(x) \exp(r(x)/\beta)\, dx$ is the normalizing constant.
Since $\mathrm{supp}(p^*) = \mathrm{supp}(p_{\mathrm{data}})$, $p^*$ is a reward-tilted version of $p_{\mathrm{data}}$: it assigns higher mass to high-reward regions within the data distribution.
We cannot evaluate $Z$ or draw samples from $p^*$ directly.
Instead, we approximate $\mathbb{E}_{x \sim p^*}[\nabla_\theta \log p_\theta(x)]$ with a practical \textbf{mixture} of two accessible sample sources that together cover the support of $p^*$:
\begin{itemize}
    \item \textbf{Offline anchor} ($x \sim p_{\mathrm{data}}$): covers the support of the data distribution, ensuring the model retains general generation quality.
    \item \textbf{Online exploration} ($\tilde{x} \sim p_\theta$): covers high-reward regions; samples are weighted by the group-relative advantage $A(\tilde{x}) = r(\tilde{x})/\beta - \bar{r}$.
\end{itemize}
This yields the mixture objective:
\begin{equation}
\label{eq:app_mixture}
\mathcal{J}(\theta) \approx \underbrace{\mathbb{E}_{\tilde{x} \sim p_\theta}\!\left[\exp\!\left(\frac{A(\tilde{x})}{\beta}\right) \log p_\theta(\tilde{x})\right]}_{\text{Online Exploration}} + \underbrace{\mathbb{E}_{x \sim p_{\mathrm{data}}}[\log p_\theta(x)]}_{\text{Offline Anchor}}.
\end{equation}

\subsubsection{Diffusion ELBO Approximation.}
\label{subsubsec:app_elbo}

The log-likelihood $\log p_\theta(x_0)$ is intractable for diffusion models.
We derive its gradient approximation from the DDPM variational lower bound (ELBO)~\cite{hoddpm}.

\textbf{ELBO decomposition.}
The evidence lower bound decomposes as:
\begin{align}
\label{eq:app_vlb}
\mathcal{L}_{\text{VLB}} &= \underbrace{D_{\mathrm{KL}}\!\left(q(x_T|x_0)\,\|\,p(x_T)\right)}_{L_T,\;\text{const w.r.t.}\;\theta} - \underbrace{\log p_\theta(x_0|x_1)}_{L_0} \nonumber \\
&\quad + \sum_{t=2}^{T} \underbrace{D_{\mathrm{KL}}\!\left(q(x_{t-1}|x_t,x_0)\,\|\,p_\theta(x_{t-1}|x_t)\right)}_{L_{t-1}},
\end{align}
where $\log p_\theta(x_0) \geq -\mathcal{L}_{\text{VLB}}$.
Using the noise-prediction reparameterization $\boldsymbol{\mu}_\theta(x_t,t) = \frac{1}{\sqrt{\alpha_t}}\!\left(x_t - \frac{\beta_t}{\sqrt{1-\bar\alpha_t}}\boldsymbol{\epsilon}_\theta(x_t,t)\right)$ and the closed-form posterior $q(x_{t-1}|x_t,x_0)$~\cite{hoddpm}, each denoising term reduces to a weighted MSE in noise space:
\begin{equation}
\label{eq:app_denoising_kl}
L_{t-1} = \frac{\beta_t^2}{2\,\sigma_t^2\,\alpha_t\,(1-\bar\alpha_t)}\,
           \bigl\|\boldsymbol{\epsilon} - \boldsymbol{\epsilon}_\theta(x_t, t)\bigr\|^2 + C.
\end{equation}
Summing over $t$ gives $\mathcal{L}_{\text{VLB}} = \mathbb{E}_{t,\boldsymbol{\epsilon}}\bigl[w_t\|\boldsymbol{\epsilon} - \boldsymbol{\epsilon}_\theta(x_t,t)\|^2\bigr] + C$ with $w_t = \frac{\beta_t^2}{2\sigma_t^2\alpha_t(1-\bar\alpha_t)}$.
Ho et al.~\cite{hoddpm} find that dropping $w_t$ (setting them to 1) yields better sample quality in practice, giving the simplified objective $\mathcal{L}_{\text{simple}} = \mathbb{E}_{t,\boldsymbol{\epsilon}}\bigl[\|\boldsymbol{\epsilon} - \boldsymbol{\epsilon}_\theta(x_t,t)\|^2\bigr]$.
Since maximizing $\log p_\theta(x_0)$ is equivalent to minimizing $\mathcal{L}_{\text{VLB}} \approx \mathcal{L}_{\text{simple}}$ (up to a sign and constant), we obtain the gradient approximation:
\begin{equation}
\label{eq:app_elbo}
\nabla_\theta \log p_\theta(x) \;\approx\; -\nabla_\theta \mathcal{L}_{\text{diff}}(x)
= -\nabla_\theta \mathbb{E}_{t,\,\boldsymbol{\epsilon}}\!\left[\|\boldsymbol{\epsilon}_\theta(x_t, t) - \boldsymbol{\epsilon}\|^2\right].
\end{equation}
This involves two approximations: (i) the ELBO lower bound, and (ii) the uniform timestep weighting; both are standard and empirically well-supported~\cite{hoddpm,nichol2021improveddenoisingdiffusionprobabilistic}.

\subsubsection{From Exponential to Linear Advantage Weights.}
\label{subsubsec:app_linearization}

Substituting Eq.~\eqref{eq:app_elbo} into the mixture objective~\eqref{eq:app_mixture} and converting from maximization to minimization yields an intermediate loss:
\begin{equation}
\label{eq:app_intermediate}
\mathcal{L}_{\text{intermediate}} = \mathbb{E}_{\tilde{x} \sim p_\theta}\!\left[
  \exp\!\left(\frac{A(\tilde{x})}{\beta}\right) \mathcal{L}_{\text{diff}}(\tilde{x})
\right]
+ \mathbb{E}_{x \sim p_{\text{data}}}\!\left[\mathcal{L}_{\text{diff}}(x)\right].
\end{equation}
The exponential weight $\exp(A/\beta)$ is standard in Advantage-Weighted Regression (AWR)~\cite{peters2007reinforcement,nair2021awacacceleratingonlinereinforcement}.
For computational stability—particularly important when advantages are clipped to a bounded range—we apply a first-order Taylor expansion:
\begin{equation}
\label{eq:app_taylor}
\exp\!\left(\frac{A(\tilde{x})}{\beta}\right) \;\approx\; 1 + \frac{A(\tilde{x})}{\beta}.
\end{equation}
The constant term contributes uniform SFT updates on generated samples.
Since the offline anchor $\mathbb{E}_{x \sim p_{\text{data}}}[\mathcal{L}_{\text{diff}}(x)]$ already provides an SFT regularization signal on \emph{real} data, we drop the redundant constant-1 contribution from generated samples and retain only the advantage-proportional term.
Absorbing the $1/\beta$ scaling into the overall loss coefficient (equivalently, viewing it as part of the learning rate), the effective online weight reduces to $A(\tilde{x})$.
With group-relative advantages $\tilde{A}^b = r_\omega(\tilde{x}^b)/\beta - \bar{r}$ (where $1/\beta$ is already incorporated), this yields the final \textbf{SHIFT} loss.

\subsubsection{Final SHIFT Loss.}

Combining the linearized online term with the offline anchor gives the SHIFT loss:
\begin{equation}
\label{eq:app_shift_loss}
\mathcal{L}_{\text{SHIFT}} = \mathbb{E}_{b,\,t}\!\left[
\underbrace{\tilde{A}^b \|\boldsymbol{\epsilon}_\theta(\tilde{x}_t, t) - \tilde{\boldsymbol{\epsilon}}\|^2}_{\text{Online RL Update}}
+ \underbrace{\|\boldsymbol{\epsilon}_\theta(x_t, t) - \boldsymbol{\epsilon}\|^2}_{\text{Offline Data Anchor}}
\right],
\end{equation}
where $\tilde{A}^b$ is the clipped group-relative advantage for the $b$-th generated sample.
When advantages are zero, SHIFT reduces to standard SFT; when advantages are large and positive, it aggressively shifts the distribution toward high-reward samples.
In practice, we clip advantages to a fixed range $[\tilde{A}_{\min}, \tilde{A}_{\max}]$ to prevent excessively large updates, analogous to the clipping in PPO~\cite{schulman2017proximalpolicyoptimizationalgorithms} and GRPO~\cite{shao2024deepseekmath}.

\section{Additional Ablation Study}
\begin{figure}[t]
  \centering
  \includegraphics[width=\linewidth]{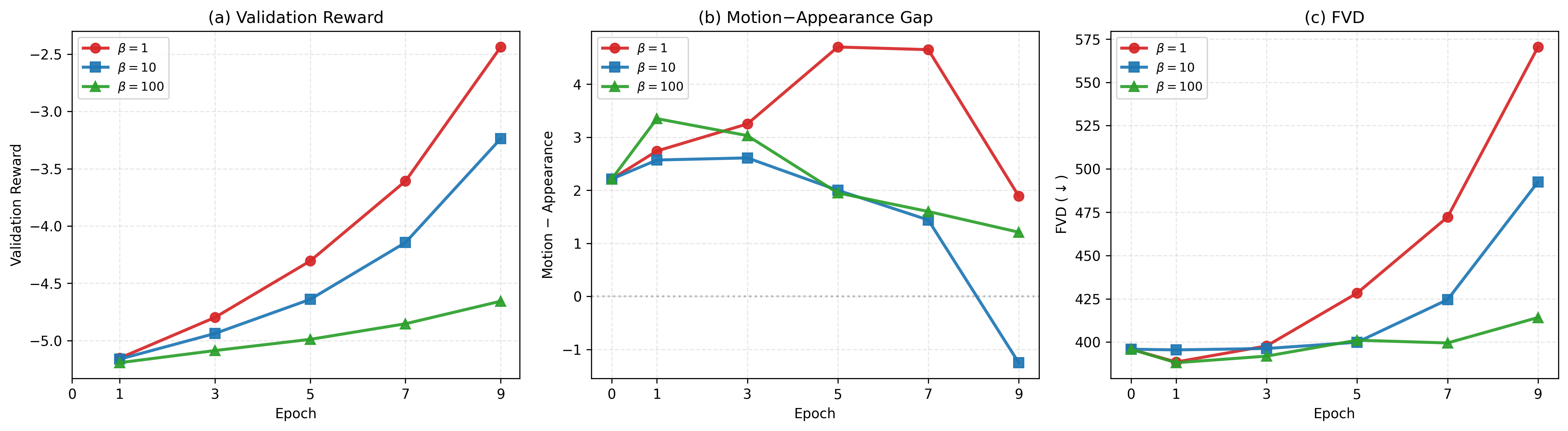}
  \caption{Effect of temperature $\beta$ on training dynamics.
  (a)~Smaller $\beta$ yields faster reward optimization.
  (b)~The motion--appearance gap reveals reward hacking: $\beta{=}1$
  peaks early then collapses as both metrics degrade; $\beta{=}10$
  turns negative; $\beta{=}100$ preserves the pre-trained balance.
  (c)~FVD monotonically increases with smaller $\beta$, confirming
  greater distributional drift under stronger RL signals.}
  \label{fig:beta_ablation}
\end{figure}

\textbf{Temperature $\beta$.}
In SHIFT, the temperature $\beta = \lambda_{\mathrm{sft}} / \lambda$
controls the balance between the SFT anchor and the advantage-weighted
RL objective, analogous to the KL coefficient in Flow-GRPO.
We ablate $\beta \in \{1, 10, 100\}$ using the IMR+NA configuration,
as shown in Figure~\ref{fig:beta_ablation}.

Smaller $\beta$ drives faster reward optimization (panel~a):
validation reward increases by 2.7 units over 9~epochs with
$\beta{=}1$, compared to 1.9 for $\beta{=}10$ and 0.5 for
$\beta{=}100$.
However, faster optimization also accelerates reward hacking.
With $\beta{=}1$, clear signs of hacking emerge after epoch~5:
the motion--appearance gap (panel~b) peaks at 4.7 and then sharply
contracts as \emph{both} appearance and motion quality degrade
simultaneously---VBench Appearance drops from 83.39 to 81.25 and
VBench Motion falls from 88.09 to 83.14, indicating that the model
has over-optimized past useful improvement.
With $\beta{=}10$, the gap turns negative by epoch~9, meaning
motion degrades below appearance under a moderate RL signal.
$\beta{=}100$ maintains a slowly shrinking positive gap, reflecting
minimal distributional change.
FVD (panel~c) corroborates this with a clean monotonic ordering:
at epoch~9, $\beta{=}1$ reaches 570 ($+44\%$ over the pre-trained
baseline), $\beta{=}10$ reaches 493 ($+24\%$), and $\beta{=}100$
reaches 414 ($+5\%$).
In summary, $\beta$ governs a continuous tradeoff between the speed
of reward learning and the degree of distributional drift;
complementary mechanisms such as Noise Alignment and Adversarial~RM
can be employed to shift this frontier.

\begin{figure}[t]
  \centering
  \includegraphics[width=\linewidth]{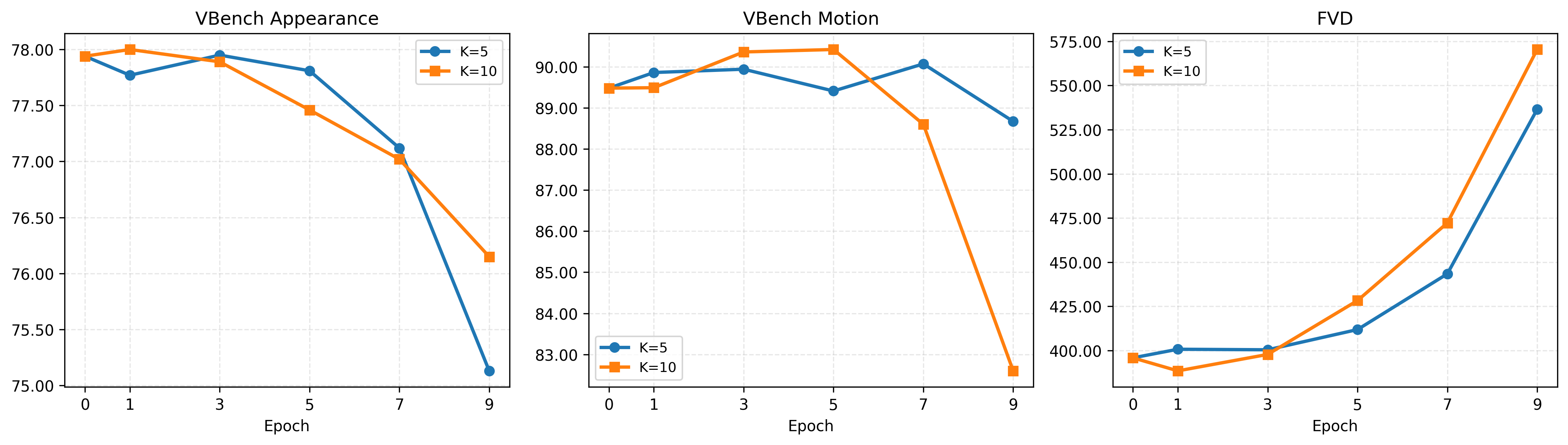}
  \caption{Ablation on buffer repeat number $K$ in SHIFT. We compare $K{=}5$ and $K{=}10$ over aligned epochs (effectively the same number of training batches). The two settings exhibit closely matched trends across all three metrics, with only minor differences throughout training.}
  \label{fig:K_ablation}
\end{figure}

\textbf{Effect of buffer passes $K$.}
As shown in Figure~\ref{fig:K_ablation}, reducing the buffer repeat number from $K{=}10$ to $K{=}5$ leads to only marginal changes in validation dynamics: VBench appearance and motion remain very close, and FVD follows a similar trajectory without a consistent qualitative shift.
These results indicate that SHIFT is robust to moderate off-policy replay, since the larger-reuse setting ($K{=}10$) remains competitive with the smaller-reuse setting ($K{=}5$).
From an efficiency perspective, however, $K{=}5$ requires more frequent buffer refresh and sampling to reach the same effective training horizon, which increases wall-clock training cost.

\textbf{IMR Encoder Architecture: ViT-Base vs.\ ResNet-34.}
We ablate the encoder backbone of the Instantaneous Motion Reward (IMR) discriminator, comparing a ViT-Base/16 encoder (${\sim}86$M parameters) against a ResNet-34 backbone (${\sim}21.5$M parameters, 2-channel input, 512$\to$256 projection, MLP head [256, 64]).
Both variants are evaluated under the IMR-only SHIFT configuration (no Noise Alignment, no Adversarial RM, no LMR), which corresponds to the A0 setting in the main paper ablation.
Results at epoch~6 are reported in Table~\ref{tab:arch_ablation}.

\begin{table}[t]
  \centering
  \small
  \setlength{\tabcolsep}{4pt}
  \renewcommand\arraystretch{1.08}
  \caption{Ablation of IMR encoder architecture evaluated at epoch~6. Both use the IMR-only (A0) configuration.
  Higher ($\uparrow$) is better; lower ($\downarrow$) is better.
  Best is \textbf{bold}.}
  \label{tab:arch_ablation}
  \resizebox{0.9\linewidth}{!}{%
  \begin{tabular}{lccccc|c}
    \toprule
    IMR Encoder & \#Params
    & \makecell[c]{VBench\\Appearance ($\uparrow$)}
    & \makecell[c]{VBench\\Motion ($\uparrow$)}
    & \makecell[c]{VBench\\Overall ($\uparrow$)}
    & \makecell[c]{Motion\\Score ($\uparrow$)}
    & \makecell[c]{FVD ($\downarrow$)} \\
    \midrule
    ResNet-34 & 21.5M & 77.27 & \textbf{94.72} & 83.09 & \textbf{5.66} & 636.2 \\
    ViT-Base  & 86.0M & \textbf{81.62} & 91.01 & \textbf{84.75} & 4.29 & \textbf{502.6} \\
    \bottomrule
  \end{tabular}%
  }
\end{table}

ResNet-34 achieves higher raw motion metrics (+3.71 VBench Motion, +1.37 Motion Score) but at a steep cost: VBench Appearance collapses by 4.35 points and FVD rises to 636.2, indicating severe reward hacking. ResNet-34's local convolutional features are more easily exploited by the generator to inflate motion rewards via superficial texture artifacts rather than genuine dynamics. In contrast, ViT-Base processes global transport-residual maps via self-attention, providing a more holistic and robust motion representation that resists such shortcuts. As a result, ViT-Base achieves substantially better VBench Overall (84.75 vs.\ 83.09) and FVD (502.6 vs.\ 636.2) while still delivering strong motion gains, confirming it as the preferred IMR encoder architecture.
We note that this comparison conflates architecture (local convolutions vs.\ global attention) with model size (${\sim}4\times$ difference in parameters); a larger model is not inherently more robust to reward hacking, as the generator can exploit richer gradient landscapes of larger models through its optimization. The improved robustness of ViT-Base is therefore attributed primarily to its global receptive field rather than its parameter count.

\textbf{Effectiveness of Noise Alignment.}
We additionally evaluate A4 with only NA removed: Appearance drops by $-0.18$ while Motion rises by $+0.32$ (MS $+0.13$), recovering the same appearance--motion asymmetry diagnosed in A0 of Table~3 of the main paper. This confirms that even in the presence of LMR and Adv.\ RM, NA is the component responsible for anchoring the SFT regularization at the high-$\sigma$ regime that AWR exploits.

\textbf{Effectiveness of the Adversarial Reward Model.}
A3 and A4 tie on VBench Overall at the best checkpoint but diverge with extended training: by epoch~9, A3 \emph{collapses} from $84.76$ (epoch~7) to $81.90$ ($-2.9$\,pp), while A4 stays within $[84.6, 84.8]$. Correspondingly, A3's validation reward climbs monotonically from $-5.15$ to $-2.22$ ($+2.93$), the classic reward-hacking signature of a fixed RM, whereas A4's reward stays bounded. Adv.~RM is therefore essential for preventing reward-hacking-induced collapse.

\section{VLM-Based Evaluation}
\label{sec:app_vlm}

We additionally evaluate Wan2.2 variants with \textbf{VisionReward}~\cite{xu2025visionrewardfinegrainedmultidimensionalhuman}, reporting aggregate reward $R$ (64 questions: 36 Appearance / 8 Dynamic / 20 other) and macro yes-rates on Appearance and Dynamic (Table~\ref{tab:vlm_eval}). VLM confirms the SFT dynamic degradation exposed in our paper: \emph{SFT's appearance gain is bought by an $-8.27$\,pp motion collapse} ($-18\%$ relative), while \emph{FlowGRPO over-corrects}, losing Appearance ($-0.78$\,pp) and dropping $R$ below base. Among motion-targeting RL methods, SHIFT beats FlowGRPO on $R$ (paired $\Delta R\!=\!+0.022$); vs.\ SFT, SHIFT wins by $+8.2$\,pp on the 8 motion questions ($+10.2$\,pp on the 4 camera-motion Q32--Q35). DenseDPO's $R$ stalls at Base (3.566 vs.\ SHIFT's 3.571), reflecting no aggregate gain. Only SHIFT improves Appearance, lifts $R$ above Base, and keeps Dynamic statistically on par with Base ($\Delta\!=\!-0.09$\,pp). No competing method achieves this near-Pareto profile. VisionReward's semantic judges complement VBench's low-level metrics.

\begin{table}[t]
  \centering
  \small
  \setlength{\tabcolsep}{4pt}
  \renewcommand\arraystretch{1.1}
  \caption{VLM-based evaluation on Wan2.2-TI2V using VisionReward~\cite{xu2025visionrewardfinegrainedmultidimensionalhuman}. We report the aggregate reward $R$ and the macro yes-rates (\%) on the Appearance and Dynamic question groups, with $\Delta$ relative to Base. Higher ($\uparrow$) is better. Best is \textbf{bold}; second-best is \uline{underlined}.}
  \label{tab:vlm_eval}
  \resizebox{0.85\linewidth}{!}{%
  \begin{tabular}{lccc}
    \toprule
    & $R$ ($\uparrow$)
    & Appearance ($\uparrow$, $\Delta$ Base)
    & Dynamic ($\uparrow$, $\Delta$ Base) \\
    \midrule
    Base         & 3.566 & 45.30 (--) & 45.39 (--) \\
    SFT          & \textbf{3.752} & \textbf{50.46}\,(+5.16) & 37.12\,($-$8.27) \\
    DenseDPO     & 3.566 & \uline{45.57}\,(+0.27) & 45.27\,($-$0.12) \\
    FlowGRPO     & 3.550 & 44.52\,($-$0.78) & \textbf{46.42}\,(+1.03) \\
    SHIFT (ours) & \uline{3.571} & 45.46\,(+0.16) & \uline{45.30}\,($-$0.09) \\
    \bottomrule
  \end{tabular}%
  }
\end{table}

\section{Implementation Details}
\label{sec:app_impl}

\textbf{Model configurations.}
For SVD, we fine-tune on the DAVIS2017 dataset~\cite{pont20172017}. Since SVD factorizes spatial and temporal attention and our motion reward targets temporal dynamics, LoRA modules (rank~32, $\alpha{=}32$, Gaussian initialization) are applied only to the temporal attention layers (projections \texttt{to\_k}, \texttt{to\_q}, \texttt{to\_v}, \texttt{to\_out}). For Wan2.2-TI2V, LoRA modules with the same configuration are applied to \emph{all} attention layers. The Wan2.2 model is fine-tuned on the deformation subset of WISA-80K~\cite{wang2025wisa}, which contains videos exhibiting rare and complex physical phenomena such as melting, tearing, and fluid deformation.

\textbf{Sampling configurations.}
For SVD, we set the motion bucket ID to 127, resolution $320\times576$~\cite{jeong2025track4genteachingvideodiffusion}, sampling frame rate 14~FPS, and 24~frames per clip. All DAVIS2017 training videos are preprocessed to match this configuration. For fair comparison with DDPO-based methods (e.g., FlowGRPO), we adopt the stochastic Euler Ancestral sampler with 25~sampling steps. We note that, unlike these baselines, SHIFT does not rely on stochastic sampling.
For Wan2.2, we use the default ODE-based sampler with 25~sampling steps, generating 49-frame clips at $480\times832$ resolution and 24~FPS, with a guidance scale of 5.0.

\textbf{SHIFT training hyperparameters.}
Table~\ref{tab:impl_details} summarizes the full training configuration for both models.
In the implementation, the temperature parameter $\beta$ from Algorithm~1 in the main paper is realized as $\beta = 1 / \lambda_{\text{awr}}$, where $\lambda_{\text{awr}}$ is the scalar coefficient on the advantage-weighted term (with $\lambda_{\text{sft}}$ fixed at 1). This is mathematically equivalent to dividing the raw reward by $\beta$ (as written in Algorithm~1) and keeping the two loss terms at equal weight.
Both models use noise alignment (\texttt{match\_sigmas}): the same randomly sampled timestep $t$ and noise $\boldsymbol{\epsilon}$ are shared between the SFT and AWR forward-diffusion passes within each training step, preventing appearance degradation from mismatched noise levels.

\textbf{Adversarial reward model co-training.}
Both the IC and LC reward models are co-trained with the generator in an adversarial fashion.
The reward models are updated with a discriminator-to-generator ratio of 1:1 per buffer pass.
We maintain an Exponential Moving Average (EMA) of the reward model weights (decay $= 0.99$); the EMA copy is used to compute advantages for the generator update, which stabilizes the reward signal.
The IC reward model uses a ViT-Base encoder operating on transport residual maps produced by SEA-RAFT~\cite{wang2024sea} optical flow, with CLS-token pooling.
The LC reward model builds on CoTracker3~\cite{karaev2025cotracker3} with 1024 randomly sampled query points per video, 4 tracking iterations, and a trajectory discriminator head.
Both reward models are optimized with AdamW (lr $= 1\times10^{-5}$, $\beta_1{=}0.9$, $\beta_2{=}0.95$).

\begin{table}[t]
  \centering
  \small
  \setlength{\tabcolsep}{4pt}
  \renewcommand\arraystretch{1.12}
  \caption{SHIFT training configurations for SVD and Wan2.2.}
  \label{tab:impl_details}
  \resizebox{\linewidth}{!}{%
  \begin{tabular}{lcc}
    \toprule
    \textbf{Hyperparameter} & \textbf{SVD (1.2B)} & \textbf{Wan2.2-TI2V (5B)} \\
    \midrule
    \multicolumn{3}{l}{\textit{Model \& LoRA}} \\
    LoRA rank / $\alpha$          & 32 / 32   & 32 / 32 \\
    LoRA target layers            & Temporal attn only & All attn \\
    Training dataset              & DAVIS2017 ($\sim$4K clips) & WISA-80K deformation \\
    \midrule
    \multicolumn{3}{l}{\textit{SHIFT Loss}} \\
    $\lambda_{\text{awr}}$ (= $1/\beta$) & 0.01  & 0.01 \\
    $\lambda_{\text{sft}}$               & 1.0   & 1.0  \\
    Temperature $\beta$                  & 100   & 100  \\
    Advantage clipping                   & $[-10, 10]$ & $[-10, 10]$ \\
    Advantage std normalization          & No & No \\
    Noise alignment (\texttt{match\_sigmas}) & Yes & Yes \\
    \midrule
    \multicolumn{3}{l}{\textit{On-Policy Buffer}} \\
    Buffer size                   & 640       & 512 \\
    Rollouts per prompt ($P$)     & 8         & 8 \\
    Passes per buffer ($K$)       & 10        & 10 \\
    \midrule
    \multicolumn{3}{l}{\textit{Sampling (Rollout Generation)}} \\
    Sampler                       & Euler Ancestral & ODE \\
    Inference steps               & 25        & 25 \\
    CFG scale                     & [1.0, 3.0]  & 5.0 \\
    Noise aug strength            & 0.02      & --- \\
    Resolution ($T{\times}H{\times}W$)   & $24{\times}320{\times}576$ & $49{\times}480{\times}832$ \\
    Frame rate                    & 14 FPS    & 24 FPS \\
    \midrule
    \multicolumn{3}{l}{\textit{Optimization}} \\
    Optimizer                     & AdamW     & AdamW \\
    Learning rate                 & $1{\times}10^{-5}$ & $1{\times}10^{-5}$ \\
    $(\beta_1, \beta_2)$          & (0.9, 0.95) & (0.9, 0.95) \\
    Weight decay                  & 0.01      & 0.0 \\
    Gradient clipping (max norm)  & 1.0       & 1.0 \\
    Warmup steps                  & 20        & 20 \\
    Batch size (per GPU)          & 5         & 2 \\
    \midrule
    \multicolumn{3}{l}{\textit{Hardware}} \\
    GPUs                          & 64 (8$\times$8) & 128 (16$\times$8) \\
    Precision                     & BF16      & BF16 \\
    Distributed strategy          & DeepSpeed ZeRO-2 & DeepSpeed ZeRO-2 \\
    \midrule
    \multicolumn{3}{l}{\textit{Reward Model Co-Training}} \\
    IMR architecture              & ViT-Base + SEA-RAFT & ViT-Base + SEA-RAFT \\
    LMR architecture              & CoTracker3 (1024 queries) & CoTracker3 (1024 queries) \\
    Reward model EMA decay        & 0.99      & 0.99 \\
    D-to-G update ratio           & 1:1       & 1:1 \\
    \bottomrule
  \end{tabular}%
  }
\end{table}


\clearpage
\section{Additional Qualitative Results}
\label{sec:app_qualitative}

\subsection{SVD Additional Results}

We provide additional qualitative comparisons for the SVD fine-tuning experiments on VBench-I2V Standard test prompts. Each example shows sampled frames from the base SVD model and different fine-tuned variants. Across diverse scenes, SHIFT consistently produces stronger motion dynamics while preserving visual quality, whereas SFT tends to reduce motion and other RL baselines may introduce artifacts.

\begin{figure}[t]
  \centering
  \begin{minipage}[t]{0.48\linewidth}
    \centering
    \includegraphics[width=\linewidth]{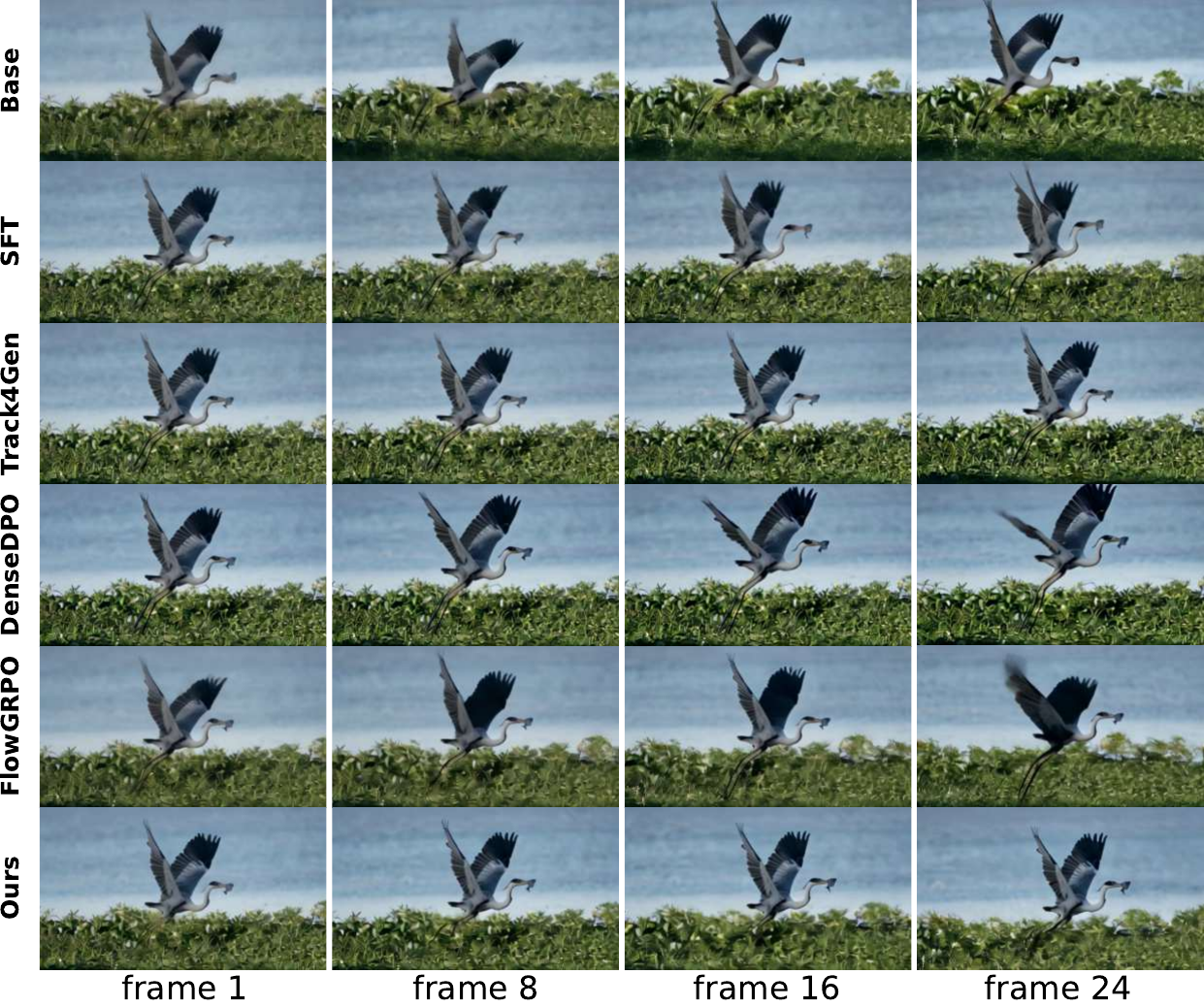}
  \end{minipage}
  \hfill
  \begin{minipage}[t]{0.48\linewidth}
    \centering
    \includegraphics[width=\linewidth]{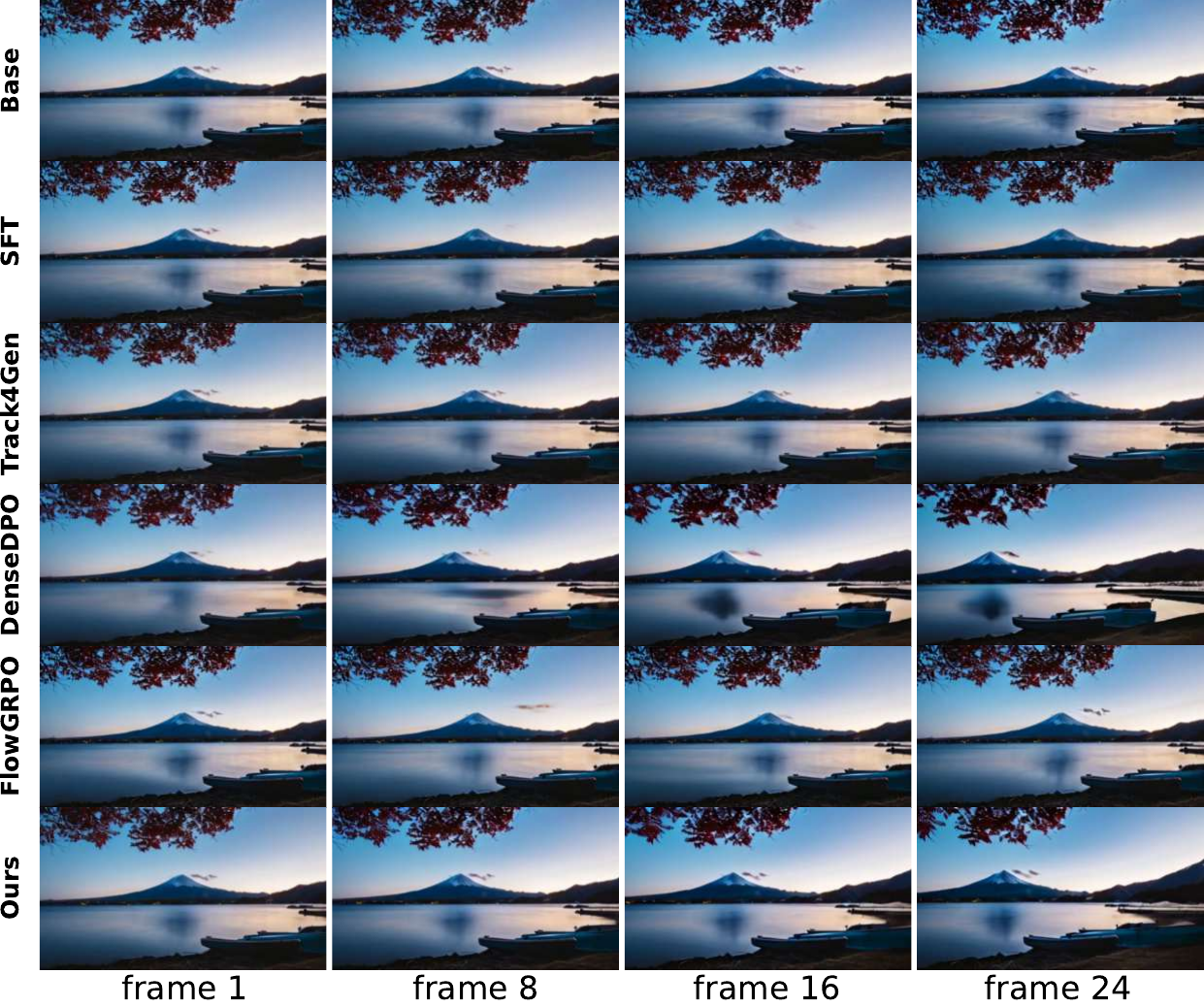}
  \end{minipage}
  \caption{Additional SVD qualitative comparisons (1/4). Left: ``a bird with a fish in its beak flying over a field''. Right: ``a boat sits on the shore of a lake with Mt Fuji in the background''.}
  \label{fig:app_svd_qual_1}
\end{figure}

\begin{figure}[t]
  \centering
  \begin{minipage}[t]{0.48\linewidth}
    \centering
    \includegraphics[width=\linewidth]{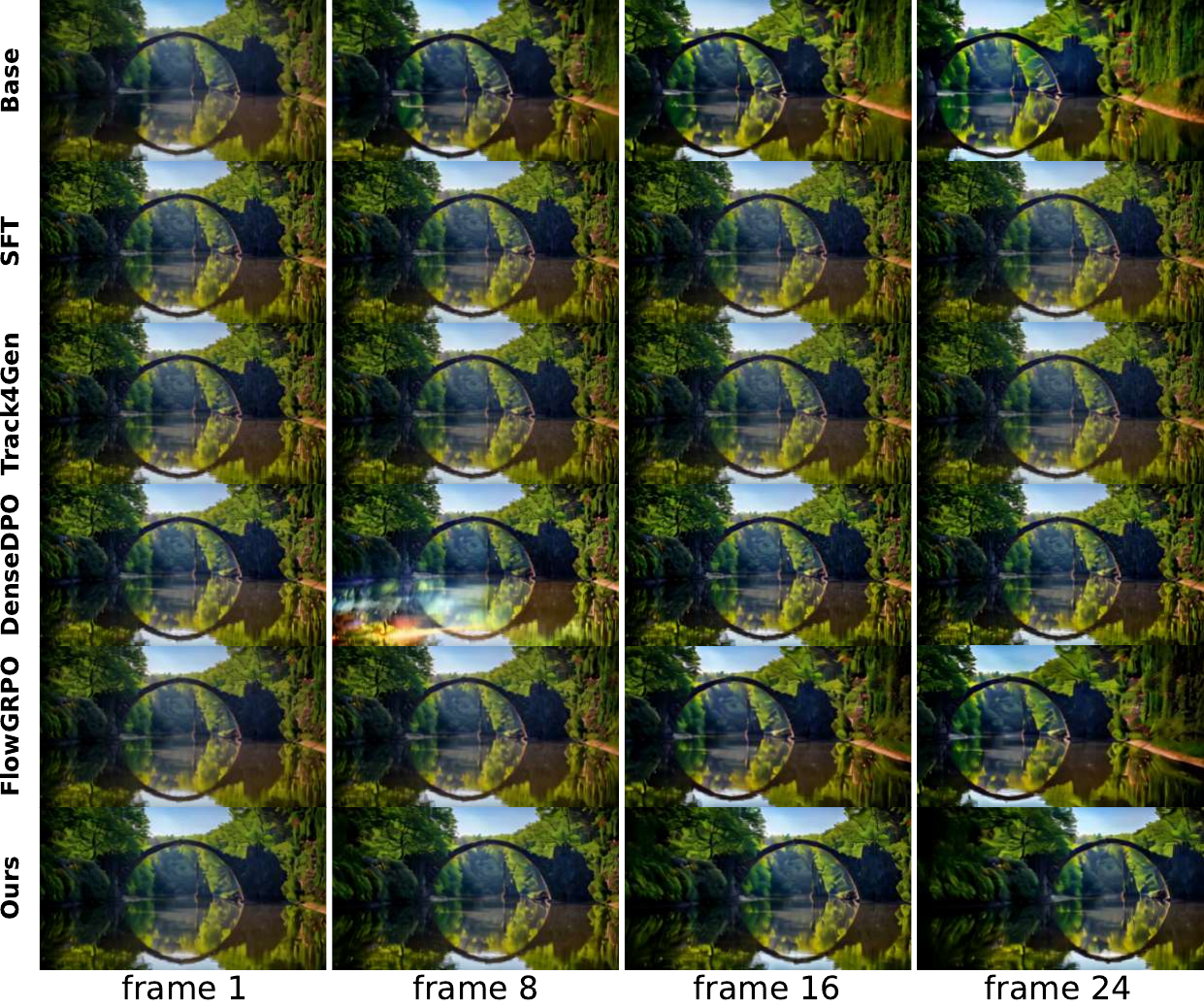}
  \end{minipage}
  \hfill
  \begin{minipage}[t]{0.48\linewidth}
    \centering
    \includegraphics[width=\linewidth]{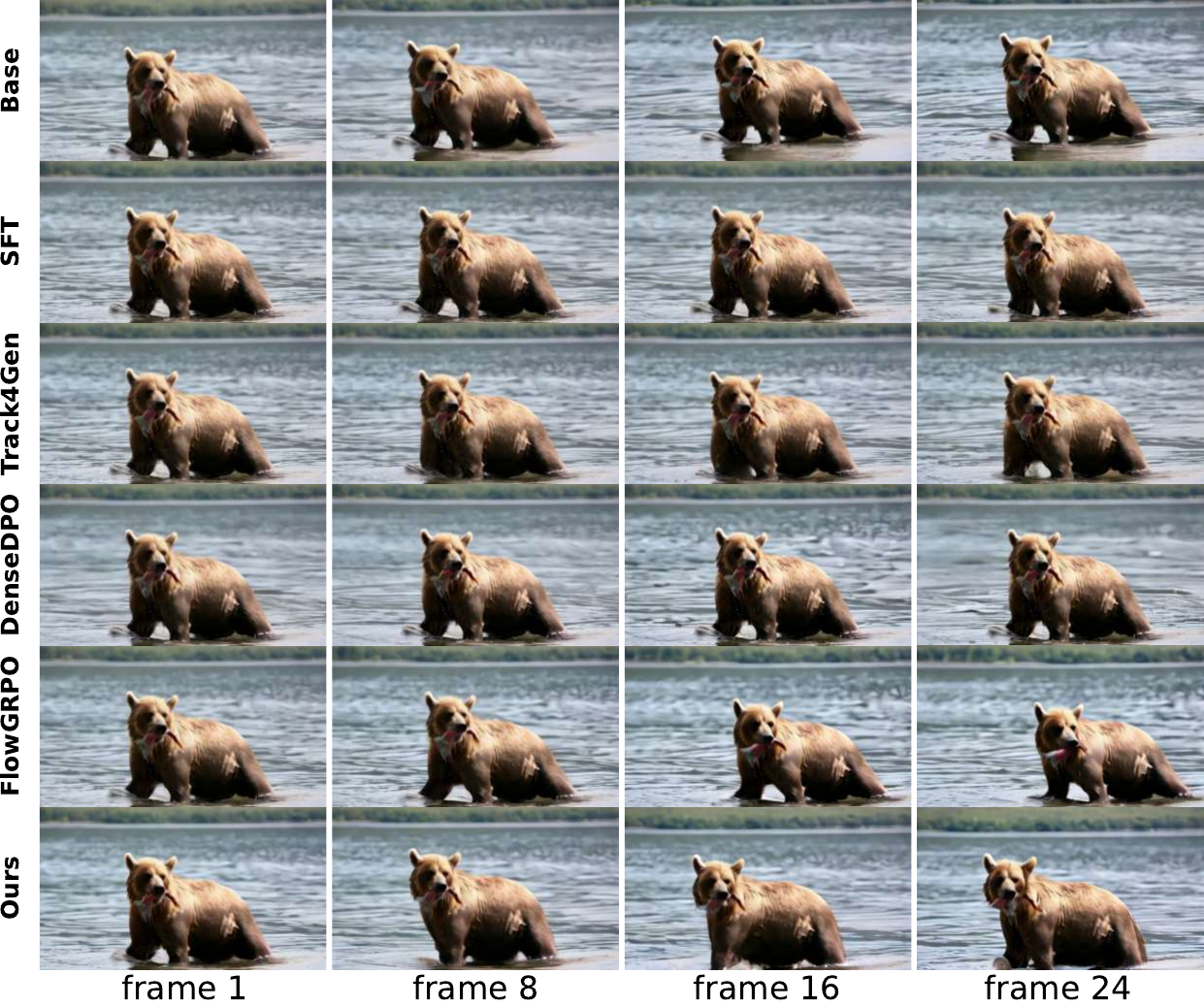}
  \end{minipage}
  \caption{Additional SVD qualitative comparisons (2/4). Left: ``a bridge that is in the middle of a river''. Right: ``a brown bear in the water with a fish in its mouth''.}
  \label{fig:app_svd_qual_2}
\end{figure}

\begin{figure}[t]
  \centering
  \begin{minipage}[t]{0.48\linewidth}
    \centering
    \includegraphics[width=\linewidth]{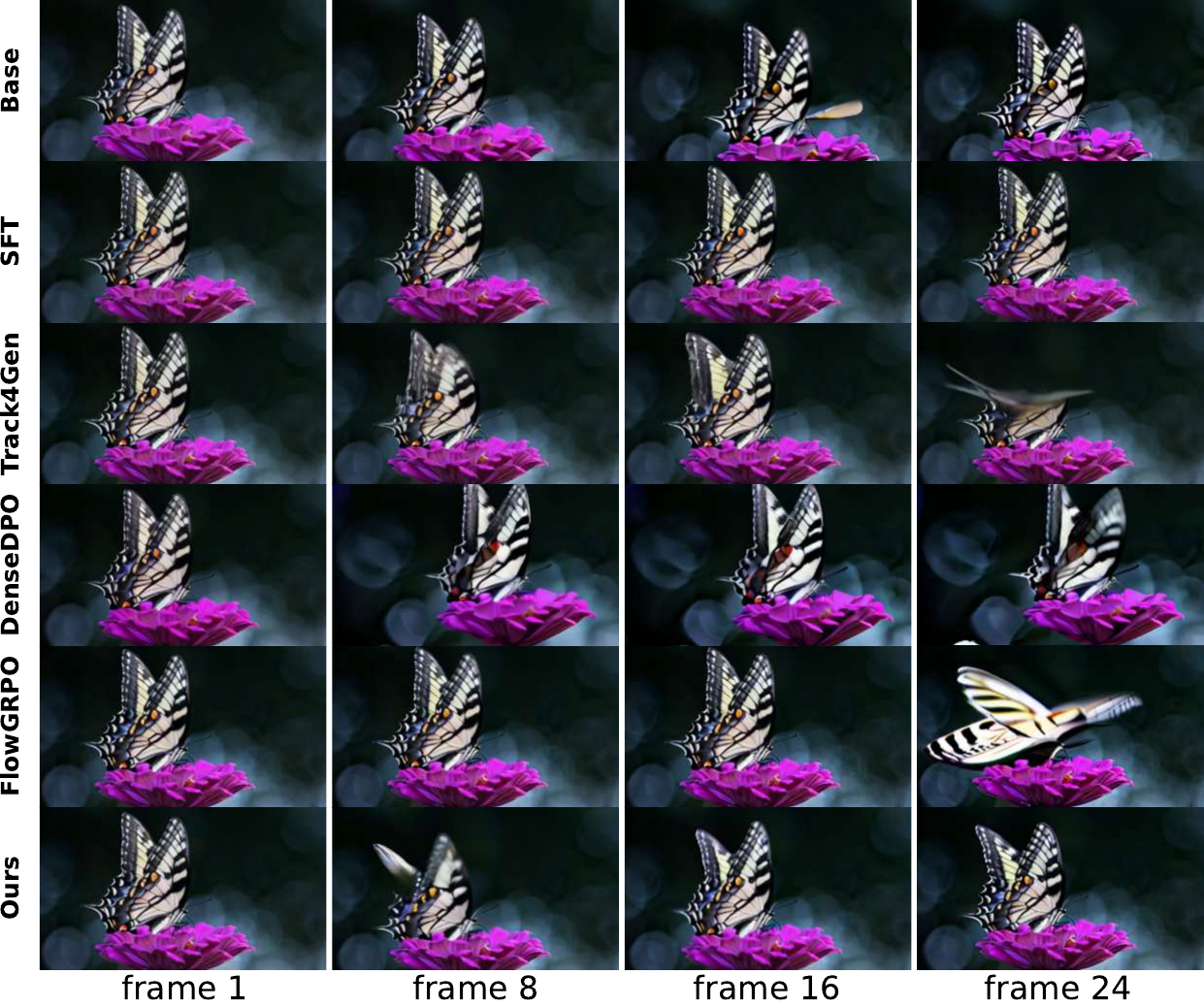}
  \end{minipage}
  \hfill
  \begin{minipage}[t]{0.48\linewidth}
    \centering
    \includegraphics[width=\linewidth]{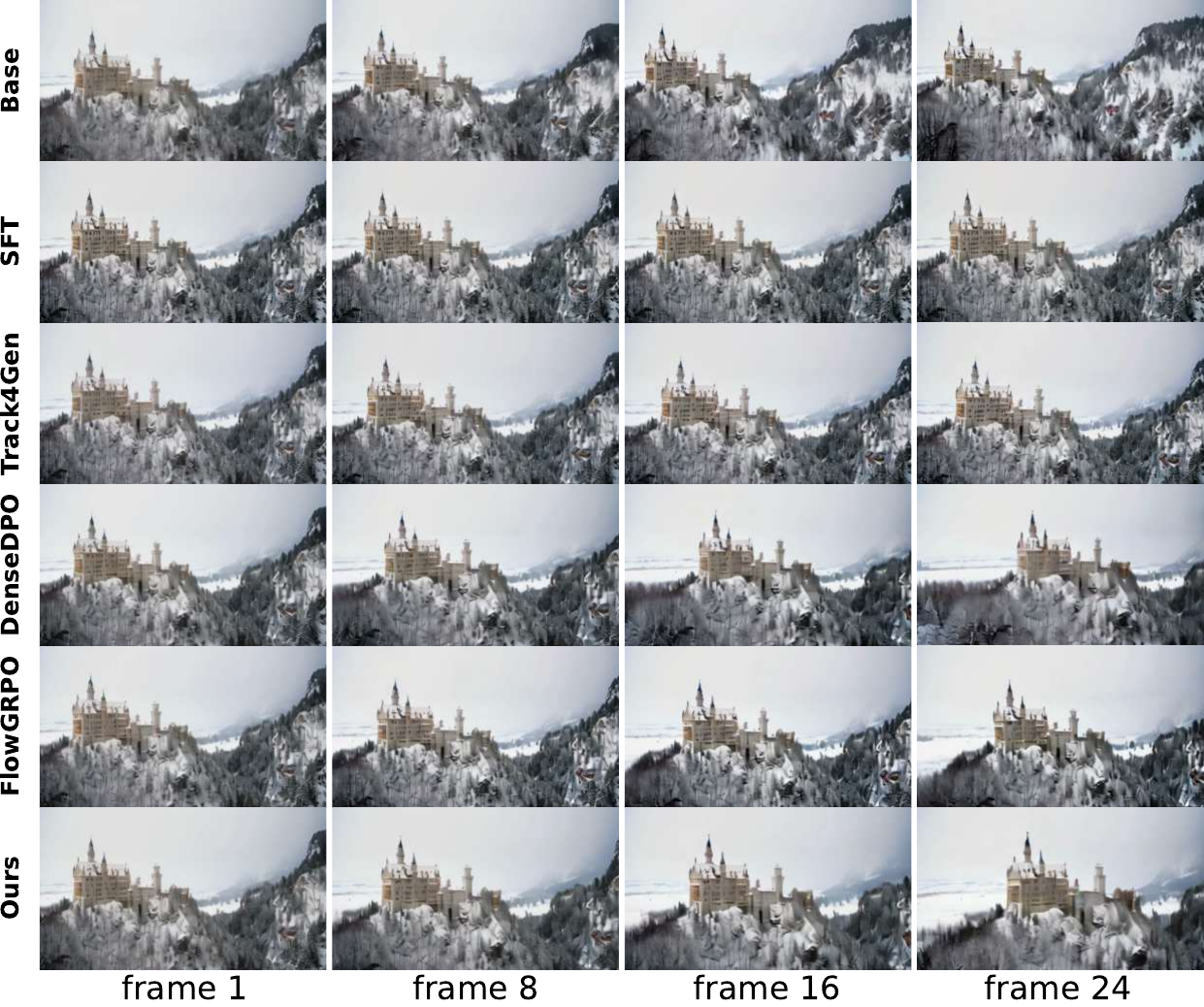}
  \end{minipage}
  \caption{Additional SVD qualitative comparisons (3/4). Left: ``a butterfly sits on top of a purple flower''. Right: ``a castle on top of a hill covered in snow''.}
  \label{fig:app_svd_qual_3}
\end{figure}

\begin{figure}[t]
  \centering
  \begin{minipage}[t]{0.48\linewidth}
    \centering
    \includegraphics[width=\linewidth]{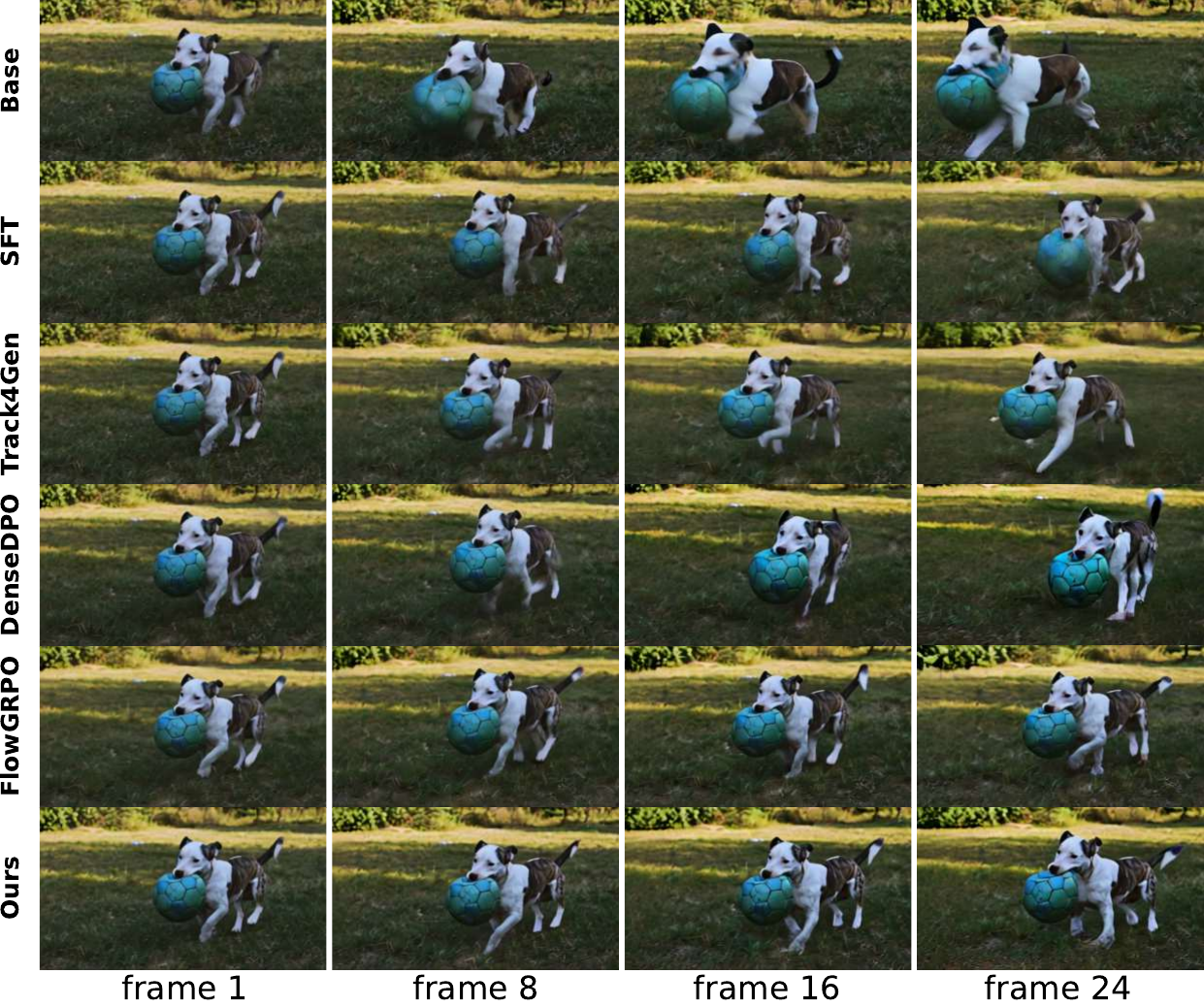}
  \end{minipage}
  \hfill
  \begin{minipage}[t]{0.48\linewidth}
    \centering
    \includegraphics[width=\linewidth]{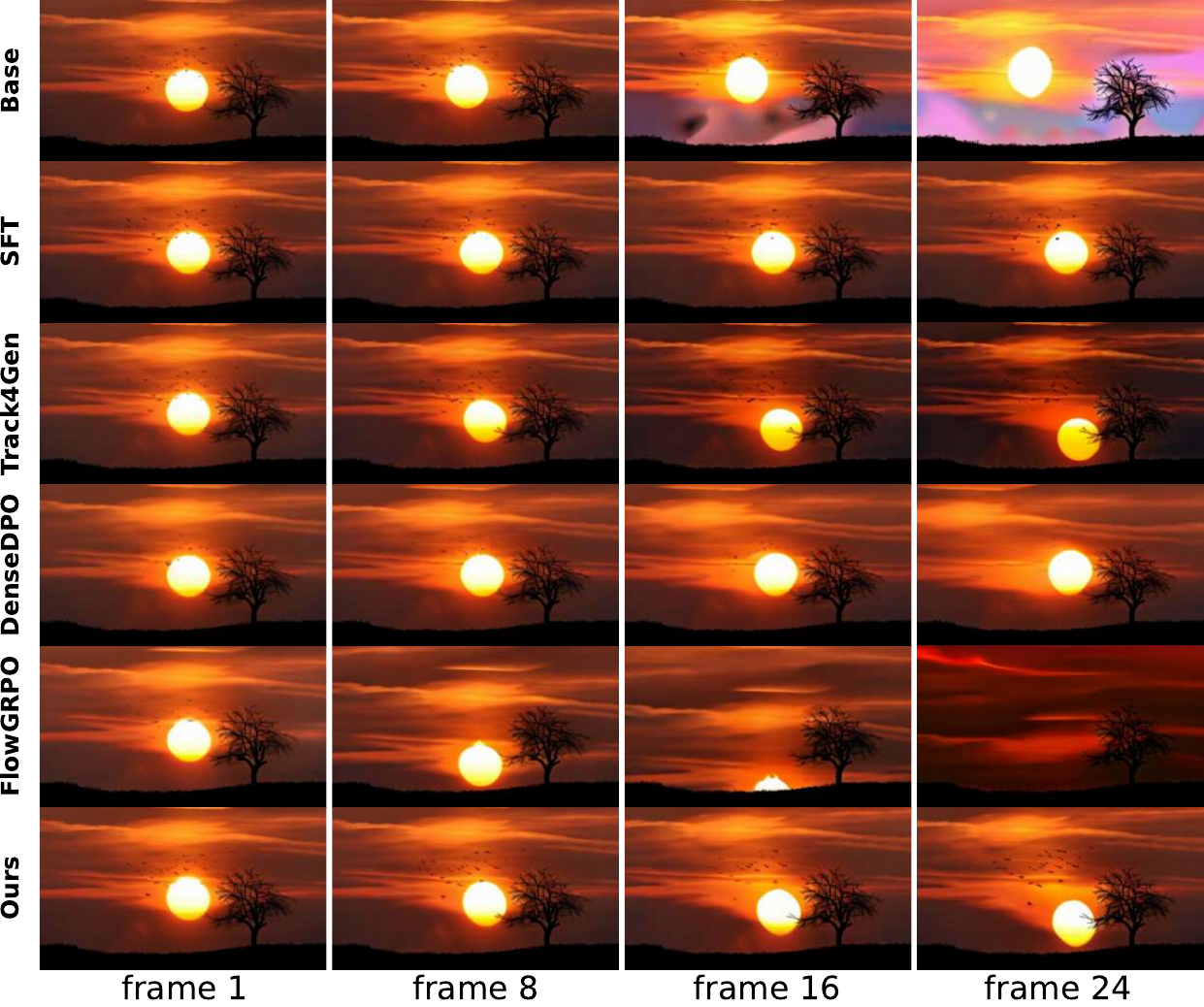}
  \end{minipage}
  \caption{Additional SVD qualitative comparisons (4/4). Left: ``a dog carrying a soccer ball in its mouth''. Right: ``a flock of birds flying over a tree at sunset''.}
  \label{fig:app_svd_qual_4}
\end{figure}

\subsection{Wan2.2-TI2V Additional Results}

We provide additional qualitative comparisons for the Wan2.2-TI2V fine-tuning experiments on the WISA-80K deformation validation set. Each example shows sampled frames from the base Wan2.2-TI2V model and fine-tuned variants. SHIFT consistently produces stronger motion dynamics than SFT while achieving better physical fidelity than the base model, particularly for complex deformation phenomena such as melting, stretching, and fluid motion.

\begin{figure}[t]
  \centering
  \begin{minipage}[t]{0.48\linewidth}
    \centering
    \includegraphics[width=\linewidth]{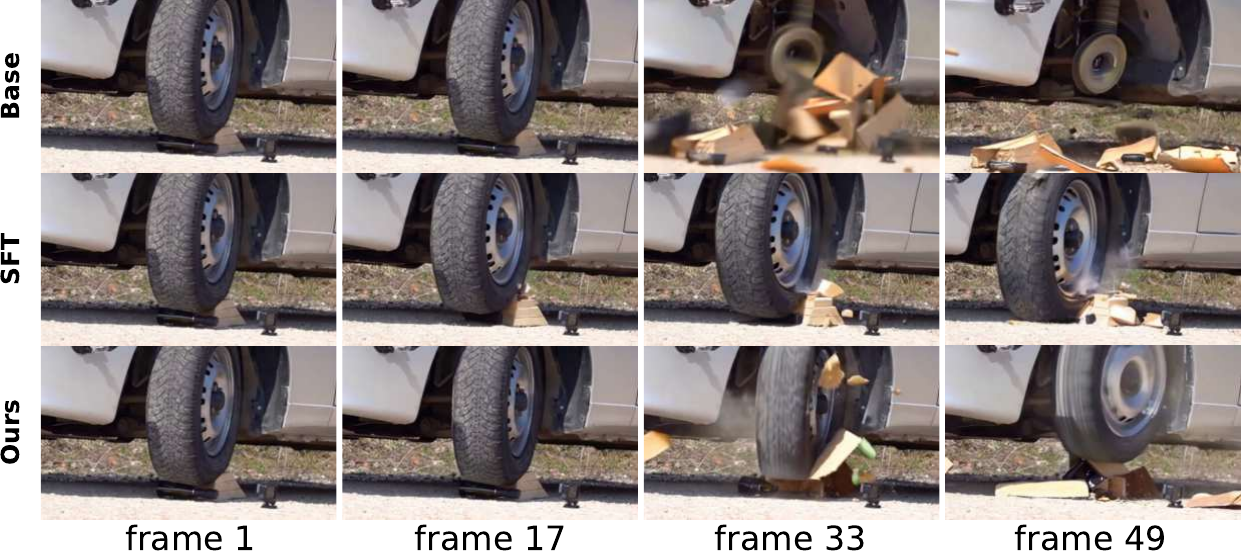}
  \end{minipage}
  \hfill
  \begin{minipage}[t]{0.48\linewidth}
    \centering
    \includegraphics[width=\linewidth]{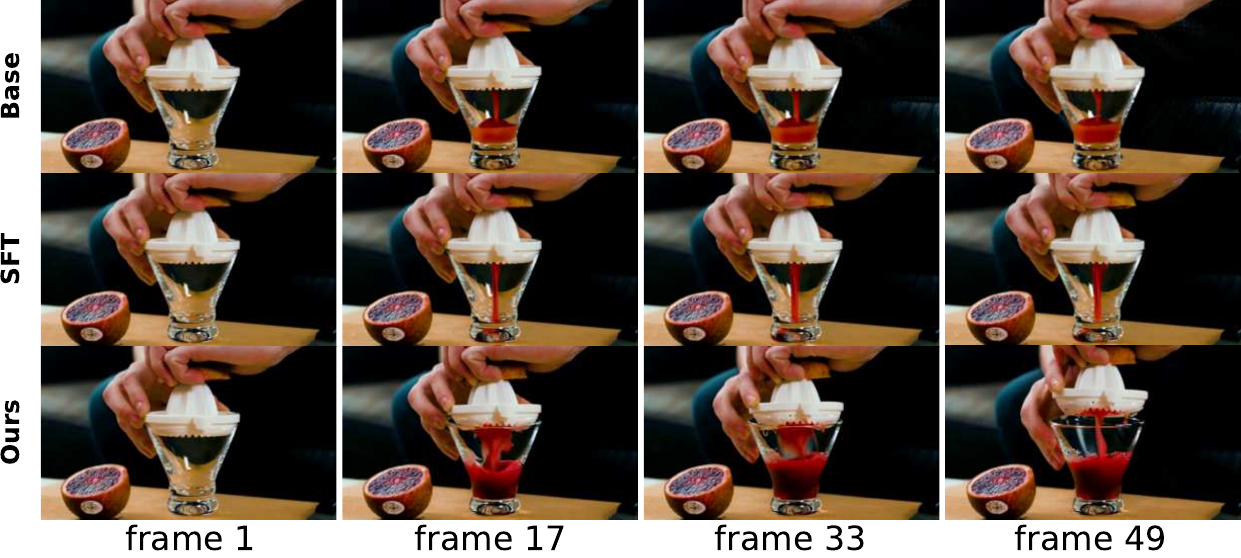}
  \end{minipage}
  \caption{Additional Wan2.2-TI2V qualitative comparisons (1/3).}
  \label{fig:app_wan_qual_1}
\end{figure}

\begin{figure}[t]
  \centering
  \begin{minipage}[t]{0.48\linewidth}
    \centering
    \includegraphics[width=\linewidth]{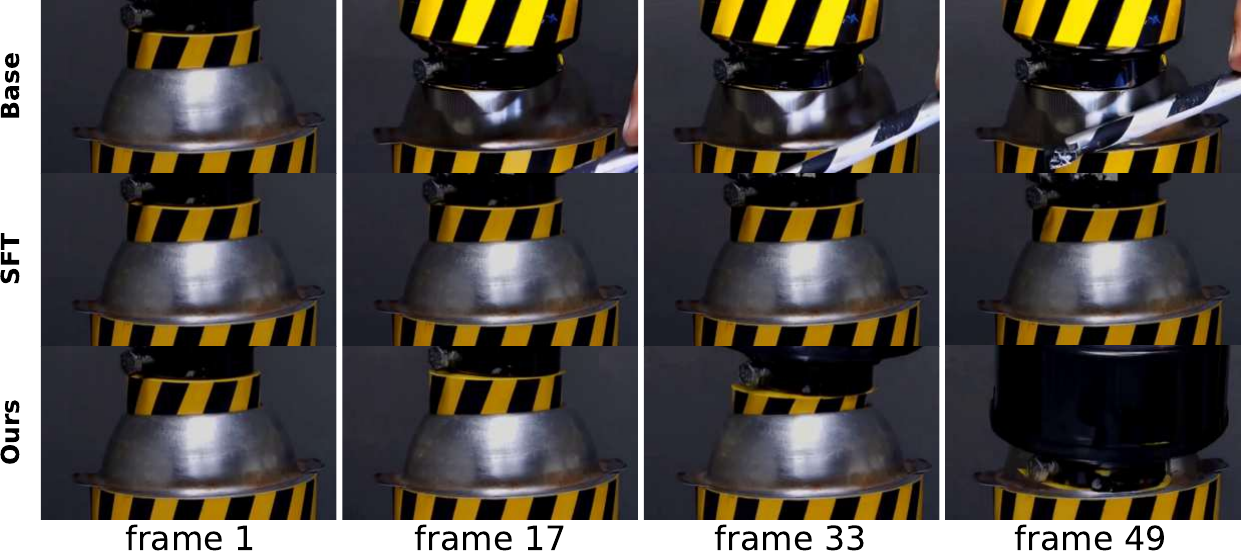}
  \end{minipage}
  \hfill
  \begin{minipage}[t]{0.48\linewidth}
    \centering
    \includegraphics[width=\linewidth]{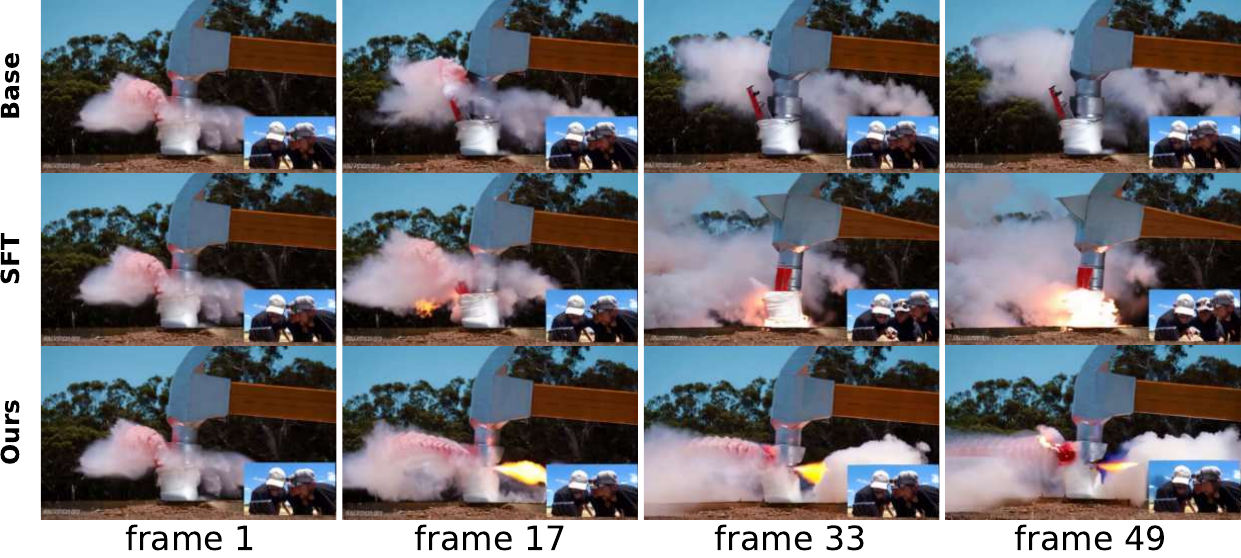}
  \end{minipage}
  \caption{Additional Wan2.2-TI2V qualitative comparisons (2/3).}
  \label{fig:app_wan_qual_2}
\end{figure}

\begin{figure}[t]
  \centering
  \begin{minipage}[t]{0.48\linewidth}
    \centering
    \includegraphics[width=\linewidth]{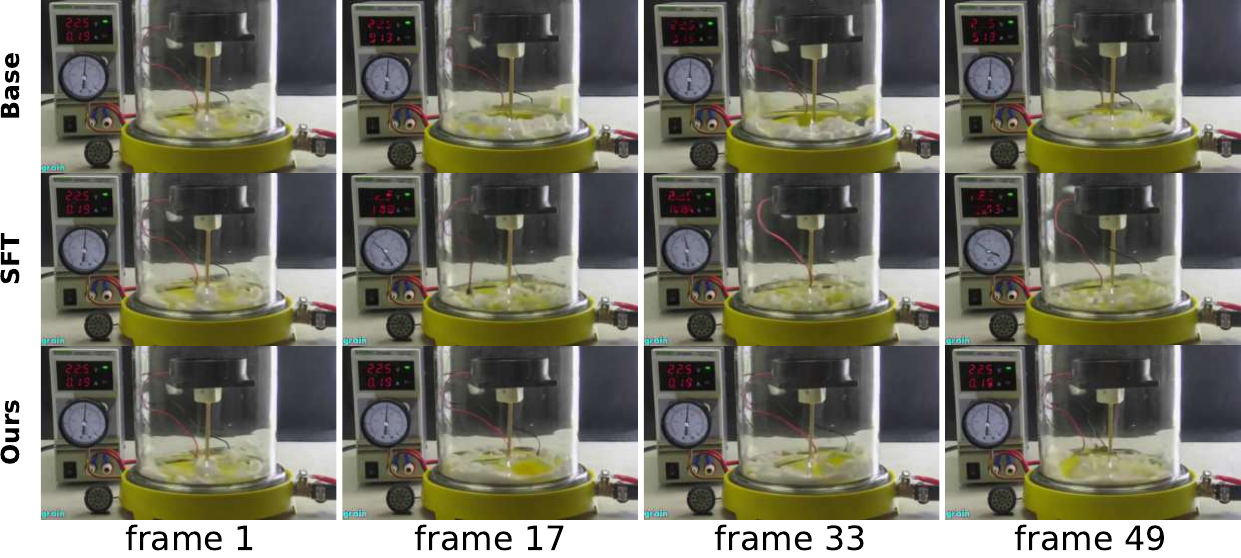}
  \end{minipage}
  \hfill
  \begin{minipage}[t]{0.48\linewidth}
    \centering
    \includegraphics[width=\linewidth]{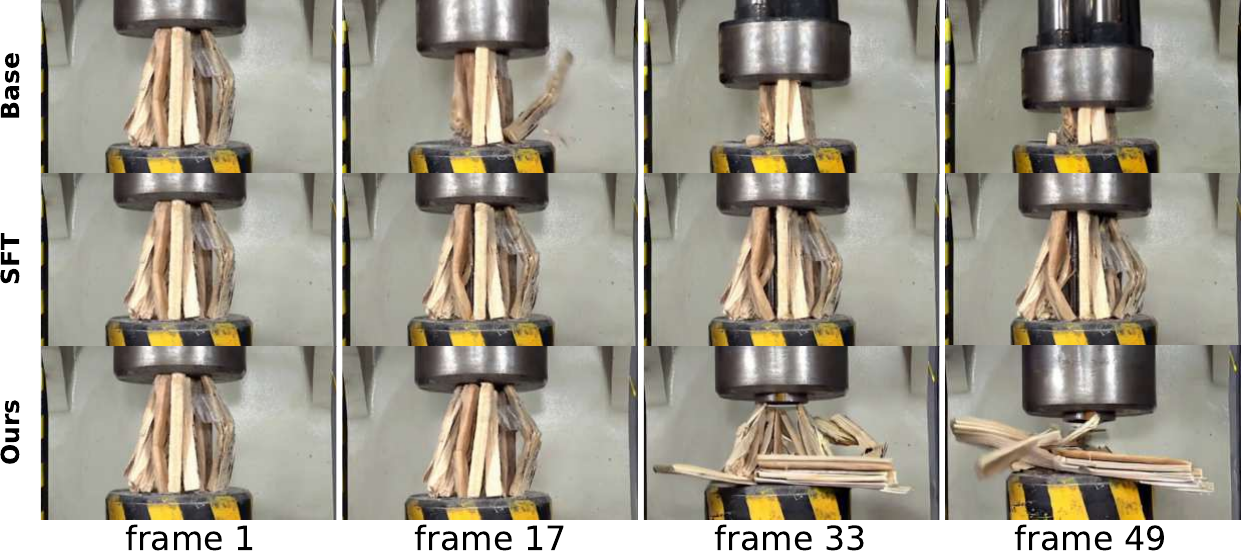}
  \end{minipage}
  \caption{Additional Wan2.2-TI2V qualitative comparisons (3/3).}
  \label{fig:app_wan_qual_3}
\end{figure}

\end{document}